\documentclass[manuscript,screen]{acmart}

\AtBeginDocument{%
  \providecommand\BibTeX{{%
    \normalfont B\kern-0.5em{\scshape i\kern-0.25em b}\kern-0.8em\TeX}}}

\setcopyright{acmcopyright}
\copyrightyear{2018}
\acmYear{2023}
\acmDOI{XXXXXXX.XXXXXXX}

\acmConference[ACM Transactions on Information Systems]{Make sure to enter the correct
  conference title from your rights confirmation emai}{}{}
\acmPrice{15.00}
\acmISBN{978-1-4503-XXXX-X/18/06}

\begin{document}

\title{(Un)likelihood Training for Interpretable Embedding}


\author{Jiaxin Wu}
\email{jiaxin.wu@my.cityu.edu.hk}
\affiliation{%
  \institution{City University of Hong Kong}
  \country{Hong Kong}
}

\author{Chong-Wah Ngo}
\email{cwngo@smu.edu.sg}
\affiliation{%
  \institution{Singapore Management University}
  \country{Singapore}
}

\author{Wing-Kwong Chan}
\email{wkchan@cityu.edu.hk}
\affiliation{%
  \institution{City University of Hong Kong}
  \country{Hong Kong}
}

\author{Zhijian Hou}
\email{zjhou3-c@my.cityu.edu.hk}
\affiliation{%
  \institution{City University of Hong Kong}
  \country{Hong Kong}
}

\renewcommand{\shortauthors}{Wu and Ngo, et al.}

\thanks{Authors’ addresses: Jiaxin Wu, Wing-Kwong Chan, and Zhijian Hou are with the Department of Computer Science, City University of Hong Kong, Tat Chee Avenue, Kowloon Tong, Hong Kong; Emails: jiaxin.wu@my.cityu.edu.hk, wkchan@cityu.edu.hk, zjhou3-c@my.cityu.edu.hk. Chong-Wah Ngo is with the School of Computing and Information Systems, Singapore Management University, 80 Stamford Road, Singapore 178902; Email:  cwngo@smu.edu.sg.}

\begin{abstract}
Cross-modal representation learning has become a new normal for bridging the semantic gap between text and visual data. Learning modality agnostic representations in a continuous latent space, however, is often treated as a black-box data-driven training process. It is well-known that the effectiveness of representation learning depends heavily on the quality and scale of training data. For video representation learning, having a complete set of labels that annotate the full spectrum of video content for training is highly difficult, if not impossible. These issues, black-box training and dataset bias, make representation learning practically challenging to be deployed for video understanding due to unexplainable and unpredictable results. In this paper, we propose two novel training objectives, likelihood and unlikelihood functions, to unroll the semantics behind embeddings while addressing the label sparsity problem in training. The likelihood training aims to interpret semantics of embeddings beyond training labels, while the unlikelihood training leverages prior knowledge for regularization to ensure semantically coherent interpretation. With both training objectives, a new encoder-decoder network, which learns interpretable cross-modal representation, is proposed for ad-hoc video search. Extensive experiments on TRECVid and MSR-VTT datasets show that the proposed network outperforms several state-of-the-art retrieval models with a statistically significant performance margin. 
\end{abstract}



\maketitle

\section{Introduction} 
Cross-modal representation learning is currently the mainstream approach for text-to-video search \cite{tpami21-dual-encoding,vse}. The approach embeds videos and user queries expressed with natural language into a common latent space for similarity measure. Compared to concept-based search \cite{ACM_TOIS_Chua_concept_based,mediamill2017,wasedaAVS2020}, this approach (aka concept-free search) bypasses the stringent challenge of selecting a small handful of concepts out of a large concept bank for video search \cite{vireoAVS2019,WasedaMeiseiSoftbank2019,ITI-CERTH2018}. The latent space is known to be more effective in contextualizing query and video features than using a sparse set of concepts to discretely model the search intent, as demonstrated in annual evaluation campaigns such as TRECVid \cite{Trecvid2021} and Video Browser Showdown \cite{VBS2021}. 

The progress of concept-free search, nevertheless, is limited by unpredictable search performance. On the one hand, query embedding is sensitive to language expression. For example, the queries {\em a bald head} and {\em a hairless head}, despite expressing the same search intention, the corresponding embeddings could be different, resulting in large search variations. On the other hand, the learning of video embedding is driven by the semantic labels associated with the training data. As the old saying goes, a picture is worth a thousand words. Fully annotating a video, which is composed of moving pictures, with thousands of words is practically impossible. Using sparsely labeled captions \cite{tgif,msr-vtt,VATEX} to train cross-modal representation will inevitably overlook the rich video content and eventually bias video embedding. The issue has recently been addressed by the dual-task model \cite{dual_task} and hybrid space \cite{tpami21-dual-encoding} by interpreting the semantic concepts underlying the embeddings while learning a latent space to embed the multi-modal features. The dual-task model, particularly, proposes likelihood training to address the issue arisen due to missing labels in learning video embedding.

\begin{figure}[t]  
\centering  
  \centering  
  \includegraphics[width=1\linewidth]{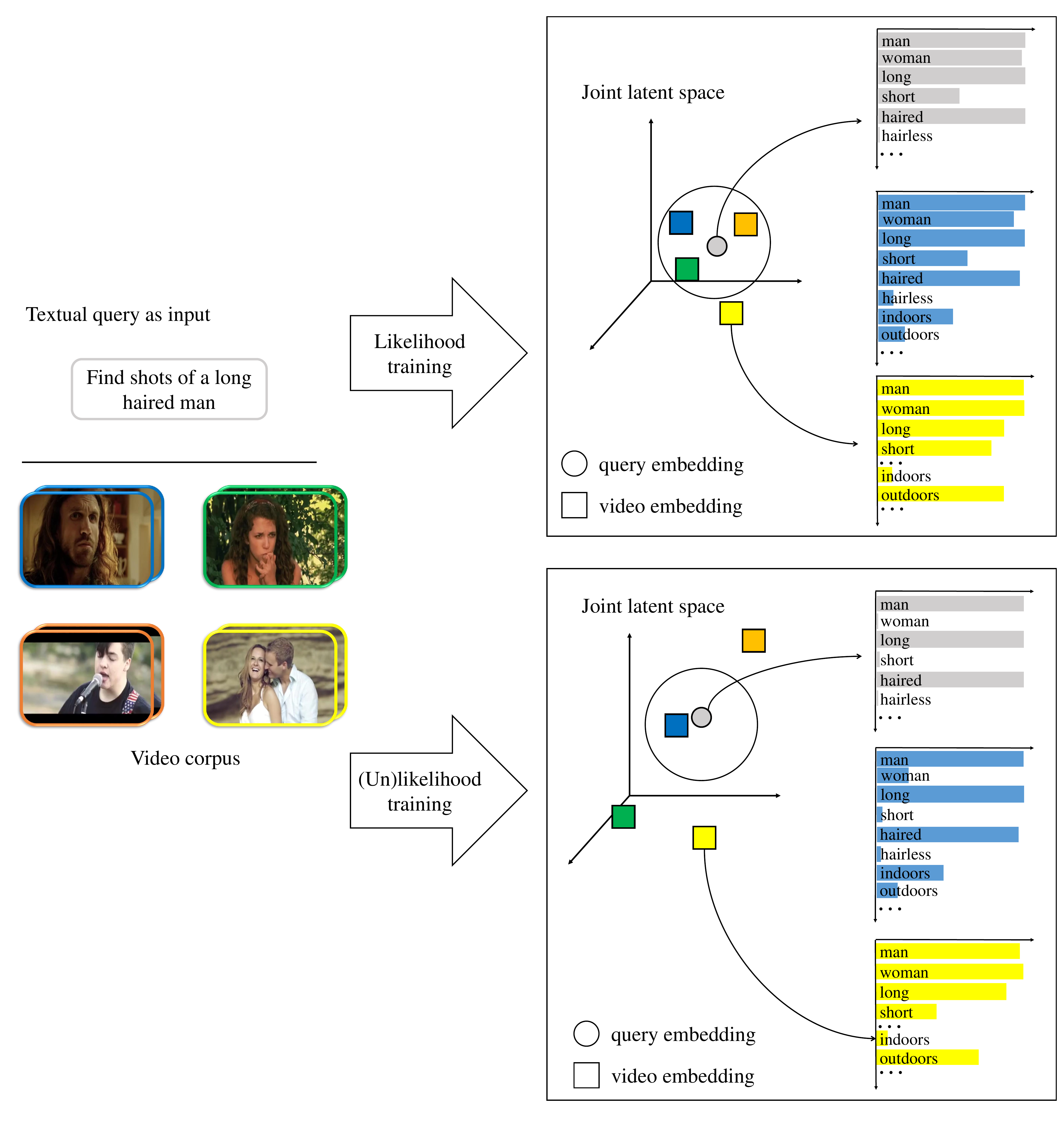}  
    \caption{A real example illustrating inconsistency in interpreting query and video embeddings. Likelihood training tends to generate interpretation that assigns high probabilities to antonym pairs (e.g., {\em man} versus {\em woman}, {\em long} versus {\em short}, {\em haired} versus {\em hairless}, {\em indoor} versus {\em outdoor}). Unlikelihood training proposed in this paper generates more semantically consistent interpretations for cross-modal embeddings. The bar charts show the probability of concept distributions interpreted from their corresponding color-coded query and video embeddings.}
    \label{fig:snapshot}
\end{figure}

This paper proposes a novel unlikelihood training along the research efforts in \cite{dual_task,tpami21-dual-encoding} to enable interpretable embedding. Specifically, likelihood training \cite{dual_task}, which leverages the concept patterns observed in the training data, does not consider semantic coherency in interpretation. The patterns, ideally, can cover relevant concepts that are not mentioned in the training caption data. However, to optimize the likelihood objective, the patterns often consist of frequent words and contextually co-occurring words that are only partially relevant or even contradict the original video or query context. Fig.~\ref{fig:snapshot} shows a user query that searches for {\em a long-haired man}, and the search result is a mix of video shots with long or short haired man and woman. By likelihood training, high probability scores are assigned to both {\em man} and {\em woman} concepts for a video shot as both concepts often co-occur in the training data. Similarly, the concepts {\em indoor} and {\em outdoor} are always interpreted together due to inconsistent labeling because these concepts are only mentioned in some training examples. Unlikelihood training addresses the inconsistency by identifying the pairs of anatomy words that could potentially lead to contradictory interpretations. With reference to Fig.~\ref{fig:snapshot}, the unlikelihood training is able to move the video shot with {\em a long-haired man} closer to the query while pushing away other near-miss video shots. Meanwhile, the interpretation is coherent by suppressing the concept scores that contradict each other and do not align with the video content. Unlikelihood training, nevertheless, cannot be trivially implemented by mutually excluding words with opposite meanings from interpretation. In the example where there are both a long-haired woman and a short-haired man in the same scene, the interpretation should equally assign high scores to the anatomy concepts. 

In this paper, we first propose unlikelihood training to regularize likelihood training and then integrate both training objectives into a unified framework for interpretable embedding. The framework learns a latent space to embed text queries and video shots while being simultaneously trained to interpret the embeddings with semantically coherent concepts. On the one hand, the likelihood objective is to encourage seamless interpretation with the prediction of relevant concepts, even if some of those concepts are not labeled in the training examples. The prediction, nevertheless, will be biased toward frequent concepts, especially those that are contextually relevant (e.g., {\em woman} and {\em man}), even if some of the concepts are not mentioned in the query or observed in a video. To safeguard the semantic consistency among the list of predicted concepts, on the other hand, the unlikelihood training regularizes the prediction by suppressing the probabilities of concepts potentially contradicting the query or video context. As representation learning and interpretation are trained end-to-end, the proposed (un)likelihood enforces faithful embedding and interpretation of query and video content, leading to higher search performance. Such training mechanism also encourages the interpretation of embedding with rare but relevant and interesting concepts, as observed in the experiments.

The main contribution of this paper is the proposal of unlikelihood training for coherent interpretations of multi-modal features embedded in a common latent space. This contribution leads to the proposal of cross-modal representation learning (Fig.~\ref{fig:framwork}) that addresses the problems of missing labels in the training data and the prediction bias due to the likelihood training objective proposed in our previous work (dual-task model \cite{dual_task}). With the newly proposed unlikelihood training (Section \ref{Sec:unlikelihood_training}), this journal version extends \cite{dual_task} by devising a new network architecture that decodes both visual and textual embeddings into concepts for coherent interpretations (Section \ref{Sec:cross_modal_video_search}). More experiments are conducted to verify the improvement over \cite{dual_task} due to these extensions (Section \ref{sec:ab_study}). The outcome is an interpretable latent space that greatly facilitates concept-free as well as concept-based searches. The rest of this paper is organized as follows. Section \ref{Sec:related_work} summarizes related work. Section \ref{Sec:interpreted_emb} first proposes the training objectives of embedding interpretation, and subsequently, Section \ref{Sec:cross_modal_video_search} presents cross-modal representation learning with (un)likelihood training. Section \ref{Sec:experiment} evaluates the proposed work on the MSR-VTT dataset and TRECVid ad-hoc video datasets across seven years of queries. Finally, Section \ref{Sec:conclusion} concludes this paper.

\section{Related Work} 
\label{Sec:related_work}
\subsection{Text-to-video search}
Text-to-video search has a long trait of history, starting from the evaluation campaign in TRECVid \cite{trecvid2006}. Given a text query, the task is to retrieve and rank videos according to their relevancy to the query. The task has evolved from video search \cite{trecvid2006}, zero-example multimedia event detection \cite{TRECVID2013}, to the more recent ad-hoc video search (AVS) \cite{George:ICMR2017_review}. Concept-based search \cite{NII2016,mediamill2017,Waseda2016,Informedia2018} was the mainstream approach before deep learning networks became popular for this task. The main idea is to index video content using large concept banks composed of a few hundred to thousands of classifiers \cite{Naphade:LSCOM,videostory,imagenet}. During search, keywords are extracted from a user query and matched against the concepts indexed for video shots. To model search intention and deal with the out-of-vocabulary problem, various research studies have been conducted for query understanding. These efforts include query expansion \cite{Waseda2018}, query-class dependent search~\cite{Kennedy:query-class-dependent-search}, ontology-based similarity measure \cite{Waseda_Meisei2017}, pseudo relevancy feedback \cite{Lu:pseudo_relevancy}, and learning from Internet videos~\cite{Informedia2018}. Nevertheless, the success of concept-based search is limited by the size of the concept bank \cite{Waseda2016} and the problem of query drifting \cite{Boer:Semantic}. Specifically, query terms are erroneously selected or expanded and improperly weighted due to a lack of query context understanding. Consequently, the search performance generally suffers from high recall and low precision.

With the rapid advance in deep learning, cross-modal embedding \cite{hgr,dualconding,tmm2021-sea,TPAMI2021_weight_metric,LAFF} gradually replaces concept-based search as the mainstream approach. The main idea is to project text queries and videos into a common latent space for representation learning. In this way, indexing is concept-free since videos are represented in high-dimensional continuous space and indexed by the embeddings extracted from deep neural networks. During search, a text query is also embedded before matching against the video embeddings for similarity ranking. Thanks to the video captioning datasets \cite{tgif,msr-vtt,VATEX,msvd}, which offer the paired data (i.e., video-caption pairs) for cross-modal representation learning, a number of networks focusing on either encoding video content\cite{dualconding} or modeling query description \cite{w2vvpp,tmm2021-sea} have been proposed. For example, dual encoding \cite{dualconding} performs multi-level embeddings of video content at global, local, and temporal levels. SEA \cite{tmm2021-sea} projects a text query into four different latent spaces for embedding. LAFF \cite{LAFF} learns weights for an optimal fusion of different embeddings learned from multi-modalities. There are other cross-modal learning methods that focus on saving storage and computation sources by hash coding the high-dimensional vision/text vectors \cite{Hu_TPAMI_CrossmodalHashing} and addressing the partially mismatched text-vision pairs problems in training \cite{HuCrossPSP_TPAMI}. Due to representation learning, concept-free approaches are relieved from the burden of concept selection. The search performance considerably surpasses concept-based search \cite{Trecvid2021}, particularly for queries that involve object attributes or inter-object relationships.

Nevertheless, concept-free search suffers from black-box training, and the learned embeddings are not interpretable. This creates the issue of search robustness. Specifically, the search result is sensitive to the formulation of a text query. The issue is first addressed in our preliminary work \cite{dual_task} by introducing a visual decoder to interpret the semantics of video embeddings. The decoder translates an embedding into a probability distribution of concepts, and the concept-based search is performed by matching the distribution with the text query terms. This work, named the dual-task model~\cite{dual_task}, enables hybrid search by leveraging the interpreted concepts in discrete space and the learned embeddings in continuous space for similarity ranking. Hybrid search basically leverages concept-based search to moderate the unpredicted performance of concept-free search due to sensitivity in query formulation. A similar idea is also explored in hybrid space learning \cite{tpami21-dual-encoding,RIVRL} later for learning latent and concept spaces. Nevertheless, different from the dual-task model~\cite{dual_task}, which learns interpretable latent space by directly attaching a decoder to the latent space, the latent and concept spaces in \cite{tpami21-dual-encoding,RIVRL} are disconnected. More specifically, the learning of two spaces is disentangled, and the coherency between latent and concept spaces is not considered in \cite{tpami21-dual-encoding,RIVRL}. Due to a lack of semantic coherency, the latent space is not interpretable compared to the dual-space model \cite{dual_task}.

\subsection{Likelihood versus unlikelihood training}
This paper extends the likelihood learning in the dual-task model \cite{dual_task} to address semantic coherency in embedding interpretation. Using video-captions pairs for learning cross-modal representation, likelihood learning allows the prediction of concepts not mentioned in the captions. As video captions only partially describe video content, this learning strategy effectively alleviates the adverse effect of missing labels. Consequently, the dual-task model demonstrates a significant boost in retrieval performance using either concept-based or concept-free search. Nevertheless, likelihood learning favors the prediction of frequent concepts even if these concepts show incoherency in semantic interpretation. For instance, both {\em indoor} and {\em outdoor} could be predicted with high probabilities for a video shot. The issue is also shared in the literature on neural text generation \cite{Unlikelihood:textGeneration,unlikelihood:ACL2020}. In predicting the next target word for sentence generation, a network trained with maximum likelihood estimation (MLE) prefers frequent over rare words. Due to the likelihood objective, this flaw has resulted in word repetition \cite{Unlikelihood:textGeneration} and inconsistent dialogue \cite{unlikelihood:ACL2020}. 
Recently, the unlikelihood training proposed in \cite{Unlikelihood:textGeneration, unlikelihood:ACL2020} has been applied by \cite{motion_unlikelihood} to prune unlikely motion trajectories that are out of drivable regions for autonomous driving. 
Inspired by \cite{Unlikelihood:textGeneration,unlikelihood:ACL2020}, this paper addresses the problem of inconsistent interpretation in the dual-task model with a soft version of unlikelihood training.

In \cite{Unlikelihood:textGeneration,unlikelihood:ACL2020}, unlikelihood training moderates MLE by explicitly penalizing the decoding of the next target words predefined in a candidate set of tokens for each conversation turn. The set is trivially composed of undesirable words and n-grams that contradict or already appear in the previous dialogue context. Different from \cite{Unlikelihood:textGeneration,unlikelihood:ACL2020}, the unlikelihood training in this paper is defined in a different context for interpreting cross-modal representation. The training is to moderate multi-label classification, which involves predicting multiple concepts from an embedding. Some of the concepts to be predicted are not even labeled or mentioned in the training data of an embedding. The level of modeling complexity is higher than \cite{Unlikelihood:textGeneration,unlikelihood:ACL2020}, where MLE is regularized by unlikelihood training to predict a single word from the previous $t$ words. More importantly, the inconsistency in interpretation depends on the context of discussion. For example, the concept {\em laughing} is opposite or inconsistent with {\em crying} when the context of an expression is about a person at a moment of time. The two concepts could become coherent if the context is about two persons, one crying while the other laughing. Different from neural text generation, where the dialog state is known, the context of the discussion is not known a priori, and hence, the inconsistency in interpretation can only be softly rather than rigidly modeled as in \cite{Unlikelihood:textGeneration,unlikelihood:ACL2020}.

\section{Interpretable Embeddings}
\label{Sec:interpreted_emb}

The goal is to interpret an embedding vector with a list of concepts describing the underlying semantics and the associated contexts. We formulate the task as a multi-label classification problem. Let $x\in \mathbb{R}^{d}$ be a $d$-dimensional embedding vector. The classifier predicts a probability distribution $\hat{G}(x)=[\hat{g}_1,\hat{g}_2,\hat{g}_3,...,\hat{g}_i,...,\hat{g}_n] \in \mathbb{R}^{n}$ over $n$ concepts, where $\hat{g}_i$ indicates the significance of a concept, represented as a probability, in explaining $x$. In principle, a classifier should predict $\hat{G}(x)$ as close as possible to the ground truth $G(x)=[g_1,g_2,g_3,...,g_i,...,g_n] \in \{0,1\}^{n}$. Nevertheless, $G(x)$ is practically sparse and provides an incomplete interpretation of $x$. Specifically, some concepts in $G(x)$ are set to be 0 despite being relevant to $x$. This situation is common in applications such as video captioning. The captions associated with a video are not likely to cover the entire word space that could describe its content. Moreover, the narration of the same scene or activities of a video can be expressed in different words or in various forms of composition. Furthermore, video content is multi-perspective. Expecting sufficient captions to describe the full spectrum of video content in the training set is impractical. 

To this end, the multi-label classifier is trained not only to predict the labeled concepts in $G(x)$ but also to speculate the unlabeled concepts that are likely or unlikely to faithfully explain the underlying meaning of $x$.  We treat the prediction of concept likelihood and unlikelihood as two separate issues. The former, likelihood training, softly learns the contextually relevant concepts through observing a large number of training examples. The latter, unlikelihood training, leverages prior knowledge (i.e., WordNet~\cite{wordnet}) to prune concepts that are contextually contrary in explaining $x$ from consideration. 

\subsection{Likelihood training} 
\label{Sec:likelihood_training}

Binary cross entropy (BCE) \cite{Murphy2012} is a conventional measure for learning multi-label classification. Nevertheless, BCE treats each concept equally, where the same degree of penalty is imposed whenever either a labeled or unlabeled concept is predicted incorrectly. As $G(x)$ is sparse with a few concepts being labeled, the BCE loss will bias towards penalizing unlabeled concepts. In explaining $x$, the unlabeled concepts are merely not mentioned rather than being not relevant. Hence, directly applying BCE will suppress relevant concepts overlooked by the training examples. In principle, the likelihood loss should only maximize the correct prediction of labeled concepts while allowing the unlabeled concepts to be predicted with a light penalty. The classifier trained in this manner is expected to minimize the loss over all the training examples by reducing the potential inconsistency in prediction due to unmentioned concepts across different examples. Imagine two videos with the presence of a lady in the scene, where one video is labeled with concept {\em woman} and the other is labeled with {\em lady}. Using this principle, a classifier that predicts {\em woman} and {\em lady} for both videos will accumulate lower penalty than a classifier that predicts either {\em woman} or {\em lady}.

Based on BCE, the likelihood loss is defined in Eq.~(\ref{eq:BCEAll}) with two terms for the prediction of label and unlabeled concepts, respectively. Note the assumption that the number of positive labels (i.e., $g_i$ =1) is much less, and hence, the first term will impose higher penalty to a concept than the second term. In other words, the equation implicitly encourages the intersection of positive labels between $\hat{G}(x)$ and $G(x)$. The second term, ideally, can allow a small set of concepts (i.e., $g_i$ = 0) to be predicted as positive. The extra loss incurred per training example is expected to be compensated by reducing the loss in other examples. Specifically, when videos with similar content are labeled with different concepts, the concepts not mentioned will be predicted for these videos. The penalties imposed by these concepts, in turn, drive the classifier to resolve potential inconsistency in labeling. Hence, the classifier is optimized to minimize the overall loss across the majority of the training examples rather than over-emphasizing to minimize the loss for individual training samples. The likelihood loss is defined as:

\begin{equation}
\begin{aligned}
\label{eq:BCEAll}
L_{BCE}(\hat{G}(x),G(x)) = \lambda \frac{1}{\sum_i^n g_{i}}\sum_i^n g_{i}BCE(\hat{g}_{i},g_{i}) +(1-\lambda) \frac{1}{\sum_i^n (1-g_{i})}\sum_i^n (1-g_{i})BCE(\hat{g}_{i},g_{i}),
\end{aligned}
\end{equation}

\begin{equation}
\label{eq:BCEloss}
BCE(\hat{g}_{i},g_{i})= -[g_{i}log(\hat{g}_{i})+ (1-g_{i})log(1-\hat{g}_{i})].
\end{equation}
where $\hat{g}_i$ is the predicted probability of the concept $i$ and ${g}_i$ refers to its ground-truth label. The parameter $\lambda$ is to control the number of unmentioned concepts that are allowed to be activated. A larger value of $\lambda$ can tolerate a larger number of unmentioned concepts, but the risk of having some of these concepts being irrelevant also increases.
More details of the hyper-parameter $\lambda$ can be found in the conference version \cite{dual_task}.

\subsection{Unlikelihood training}
\label{Sec:unlikelihood_training}

Unlikelihood training aims to ensure prediction consistency within a video shot. For example, the concepts {\em indoors} and {\em outdoors} should not be predicted as positive simultaneously. Following \cite{ACL:contradiction_in_text}, we attach each concept with a list of its antonyms as exclusive (contradicting) concepts extracted from WordNet~\cite{wordnet}. With this prior knowledge, the unlikelihood loss function is then designed to impose a penalty for the prediction of antonym pairs. Let $T_i$ be the set of antonyms (e.g., a concept $t$) obtained for a concept $i$, and then the loss function is as follows:
\begin{equation}
\begin{aligned}
\label{eq:oriUL}
L_{UL}(\hat{G}(x),G(x)) = \frac{1}{\sum_i^n g_{i}}\sum_i^n -g_i \sum_{t \in {T_i}} log(1-\hat{g}_t)   
\end{aligned}
\end{equation}
Note that the equation is similar to the proposal in \cite{unlikelihood:ACL2020} for dialog conversation, except that $T_i$ is defined for each concept rather than each dialog turn in \cite{unlikelihood:ACL2020}. This equation, nevertheless, is only applicable for the strictly exclusive terms such as time (e.g., {\em day} versus {\em night}) and scene (e.g., {\em indoor} versus {\em outdoor}) related concepts. In general, the scope of an antonym depends on the context that it refers to. For example, a video shot with the presence of {\em a woman} does not exclude the presence of {\em a man} in the same scene. Similarly, an action (e.g., {\em standing}) should only be associated with the subject taking the action and not other subjects appearing in a video. For instance, it is possible to have a video shot with both {\em a man is standing} and {\em a woman is sitting down}, and the concepts {\em standing} and {\em sitting} are not mutually exclusive. In general, we can classify an antonym pair as globally or locally exclusive concepts, depending on whether the presence of a concept can globally rule out its antonym from being observed in a video shot. The details of extracting antonym pairs from WordNet are presented in Section \ref{Sec:experiment_setting}.

To this end, we modify Eq.~(\ref{eq:oriUL}) to take account of locally exclusive concepts as well, as follows:
\begin{equation}
\begin{aligned}
\label{eq:unlikelihood}
L_{UL}(\hat{G}(x),G(x)) =\frac{1}{\sum_i^n g_{i}}\sum_i^n -g_{i}\sum_{t\in T_i}(log(1-\hat{g}_{t}))*(1-g_{t})
\end{aligned}
\end{equation}
Specifically, the equation introduces the term ($1-g_t$) to cover both locally and globally exclusive concepts such that penalty is incurred only when exclusive concepts are not mentioned in the training examples of $x$ (i.e., $g_t$ = 0). Eq.~(\ref{eq:unlikelihood}) represents a generic form that relies on the training examples to deal with both globally and locally exclusive concepts implicitly. Note that, in case of labeling error or noise, the constraint of $g_i \neq g_t$ can be explicitly enforced for globally exclusive concepts based on prior knowledge and regardless of training examples. For locally exclusive concepts, nevertheless, Eq.~(\ref{eq:unlikelihood}) takes the risk of imposing a penalty on an anatomy counterpart that is not mentioned rather than being not relevant. The risk is counterbalanced by combining both likelihood and unlikelihood training as follows:
\begin{equation}
\begin{aligned}
\label{eq:concept_loss}
loss_{concept}(\hat{G}(x),G(x))  = L_{BCE}(\hat{G}(x),G(x)) + \alpha * L_{UL}(\hat{G}(x),G(x)) 
\end{aligned}
\end{equation}
where $\alpha$ is a trade-off parameter. In Eq.~(\ref{eq:BCEAll}), the likelihood training aims to activate the relevant concepts not being mentioned for the explanation of $x$. Meanwhile, unlikelihood training attempts to moderate the interpretation by selectively downgrading the probabilities of some activations based on the prior knowledge (i.e., $T_i$ in Eq.~(\ref{eq:unlikelihood})). The pull-and-push factor due to counterbalance in two training strategies introduces perturbations in model training. As discussed in \cite{Nicola:push-pull,Nitish:Dropout}, such pull-push perturbations can practically improve the robustness of neural networks, which is also verified in our empirical studies.

\section{Cross-modal Video Search}
\label{Sec:cross_modal_video_search}

\begin{figure*}[t]
    \centering
    \includegraphics[width=1\linewidth]{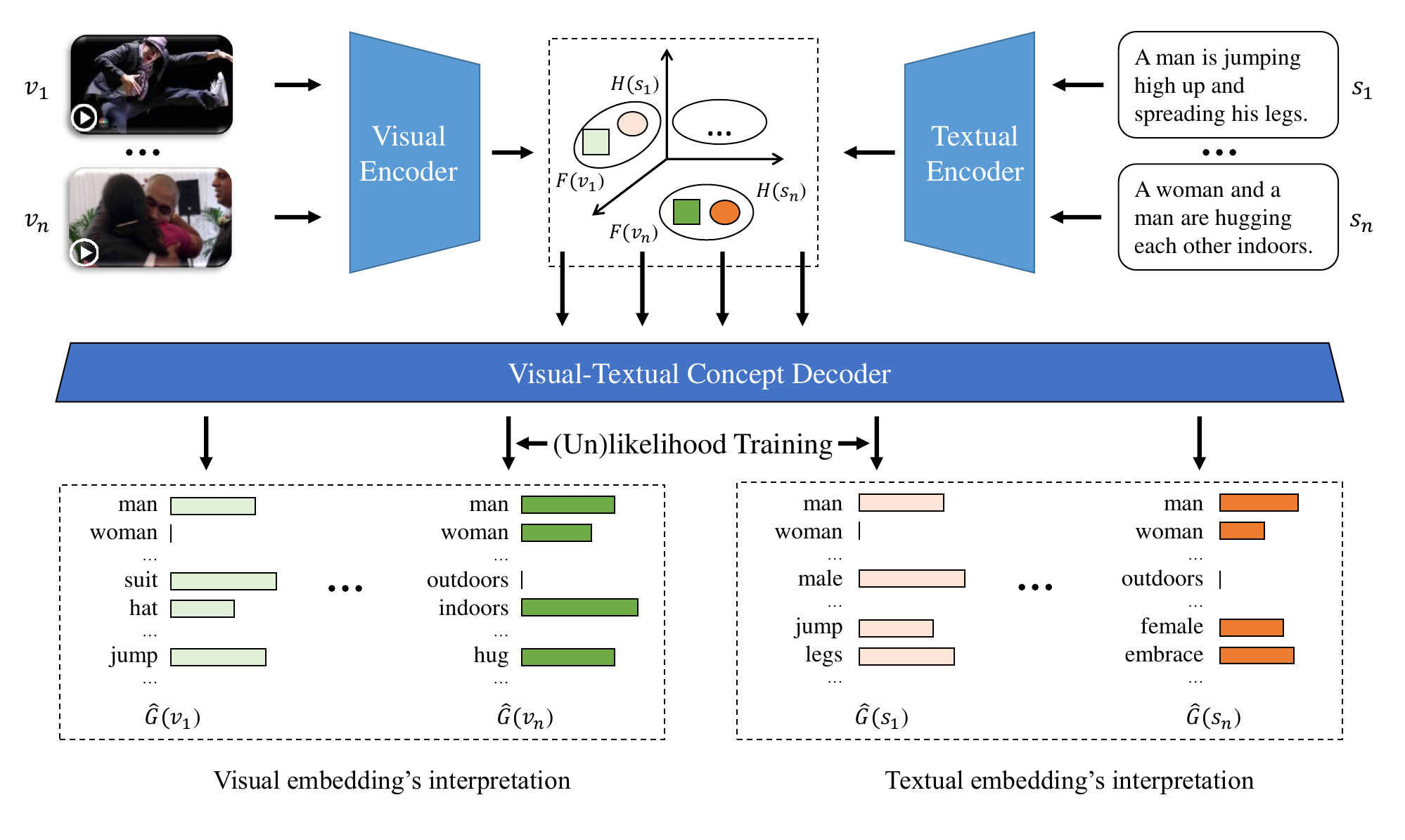}
    \caption{The proposed architecture for cross-modal video search.}
    \label{fig:framwork}
\end{figure*}

This section presents the idea of incorporating (un)likelihood training into cross-modal representation learning. The overall framework is depicted in Fig.~\ref{fig:framwork}. Similar to the existing works \cite{dualconding,tmm2021-sea,hgr}, the proposed model learns the projections to embed videos and their captions, respectively, into a cross-modal latent space. In this space, the pairs of textual and visual embeddings stay close to each other. With (un)likelihood training, the embeddings are interpreted with text-based concepts reflecting their underlying semantics and associated contexts. Moreover, the interpretations of visual and textual embeddings are trained with the same ground truth to build a connection. The model is end-to-end trained to ensure that the learned cross-modal representations are explainable with as complete and consistent as possible concept lists.  

\subsection{Search model}
\label{method_search_model}

As shown in Fig.~\ref{fig:framwork}, the network is composed of two encoders for representation learning and one decoder for embedding interpretation. The visual encoder consists of five components: pre-trained CNN networks \cite{resnet,resnext}, a biGRU network \cite{gru}, a biGRU-CNN network \cite{dualconding}, a pre-trained swin-transformer network \cite{SwinTrans}, and a pre-trained SlowFast network \cite{slowfast}. The resulting five features are concatenated to represent a video $v$ as $f(v)=[$CNNs$(v), $biGRU$(v), $biGRU-CNN$(v), $swinTrans$(v), $SlowFast$(v)]$. The textual encoder has a similar architecture, which contains bag-of-word (BoW), biGRU, and biGRU-CNN to encode a sentence $s$ with $h(s)=[$BoW$(s), $biGRU$(s), $biGRU-CNN$(s)]$. A fully connected (FC) layer and a batch normalization (BN) layer are followed to map visual/textual features into the same latent space. 
\begin{equation}
\begin{aligned}
\label{eq:visual_mapping}
F(v) = BN(\textbf{W}^{1}f(v)+b^{1})
\end{aligned}
\end{equation}
\begin{equation}
\begin{aligned}
\label{eq:textual_mapping}
H(s) = BN(\textbf{W}^{2}h(s)+b^{2})
\end{aligned}
\end{equation}
where the matrices $\textbf{W}^{1}$, $\textbf{W}^{2}$, and the vectors $b^{1}$, $b^{2}$ are weights and biases of the FC layers in the encoding networks.

Our visual-textual concept decoder is comprised of an FC layer, a BN layer, and a sigmoid function to interpret an embedding, i.e., $F(v)$ or $H(s)$, as a probability distribution of concepts. The following show the outputs for video and text embeddings:

\begin{equation}
\begin{aligned}
\label{eq:classifcation_v}
\hat{G}(v) = sigmoid(BN(\textbf{W}^{3}F(v)+b^{3}))
\end{aligned}
\end{equation}

\begin{equation}
\begin{aligned}
\label{eq:classifcation_s}
\hat{G}(s) = sigmoid(BN(\textbf{W}^{3}H(s)+b^{3}))
\end{aligned}
\end{equation}
The matrix $\textbf{W}^{3}$ and the vector $b^{3}$ are the weight and bias of the FC layer in the decoder, respectively. As a pair of visual and textual embeddings ideally stay close in the latent space, their interpreted concept lists should be at least partially overlapped. For example, a text embedding of {\em people walking on the beach} could be interpreted with the scene-related concepts such as {\em sand} and {\em ocean} observed in a video. 

\subsection{Joint embedding and interpretation}
\label{Sec:joint_embedding_and_interpretation}

The network is trained with the objectives of maximizing compatible cross-modal embeddings and consistent interpretation simultaneously. The objectives are formulated as two loss functions to characterize the latent space for its capacity in embedding and interpreting multi-modal data. The first function employs triplet loss \cite{vse} to minimize the distance between text and video embeddings in the latent space by
\begin{equation}
\begin{aligned}
\label{eq:tripletloss}
&loss_{embedding}(H(s),F(v)) =\\ &max(0,c+Sim(H(s),F(v^{\_}))-Sim(H(s),F(v))) \\
&+max(0,c+Sim(H(s^{\_}),F(v))-Sim(H(s),F(v)))
\end{aligned}
\end{equation}
where $s^{\_}$ and $v^{\_}$ are negative samples for $v$ and $s$, respectively, and $c$ represents the margin. We use cosine similarity as a measure, i.e., $Sim(s,v)=cos\_sim(H(s), F(v))$. The second function is based on (un)likelihood loss in Eq.~(\ref{eq:concept_loss}) to measure the quality of interpreting visual and text embeddings, respectively. Note that the ground truths $G(v)=G(s)$ are binary vectors capturing the words present in the captions of a video shot. Combining both types of losses, we have 
\begin{equation}
\begin{aligned}
\label{loss_matching}
loss =loss_{embedding}(H(s),F(v))+loss_{concept}(\hat{G}(v),G(v))+loss_{concept}(\hat{G}(s),G(s))
\end{aligned}
\end{equation}

\subsection{Inference}
\label{Sec:inference}

The proposed network basically enables concept-based search \cite{Waseda2016,Lu:ICMR}, embedding (aka concept-free) search \cite{w2vvpp,dualconding}, and a late fusion of both search paradigms. In concept-based search, the video shots in a dataset are indexed with concepts interpreted from their embeddings. During search, a text-based query is embedded and then interpreted with a concept list. The list is matched against the indices of a video database to produce a ranked list of search result. In embedding search, video shots are indexed with high dimensional vectors $F(v)$, which will be compared directly with query embedding for similarity ranking. The ranked lists from the concept-based and embedding searches can also be lately fused \cite{dual_task,tpami21-dual-encoding}. In the following sections, we name our method as Interpretable Text-to-Video search (ITV).

\textbf{Embedding search (ITV$_{embedding}$)}. Given a test query $q$, the textual encoder encodes $q$ as a text embedding $H(q)\in \mathbb{R}^d$. Based on the similarity of text and video embeddings, a score is computed for each video $v$:
\begin{equation}
\label{eq:score_emebdding}
Score(q,v)_{embedding}= cos\_sim(H(q),F(v)).
\end{equation}
All videos are sorted based on the embedding scores. The top-1 in the list represents the most similar video for the input query.

\textbf{Concept-based search (ITV$_{concept}$)}. The embedded query $H(q)$ is further decoded into $\hat{G}(q)\in \mathbb{R}^{n}$ and matched against $\hat{G}(v)$ indexed in the database. We consider only those concepts where $\hat{g}_i>0.99$ for matching. The reason is that in the histogram of $\hat{G}(q)$, an obvious spark is observed in the threshold of 0.99. Empirical evidence shows that those concepts are representative enough for the query. The similarity between query $q$ and video $v$ is computed as:
\begin{equation}
\label{eq:score_concept}
Score(q,v)_{concept}= \hat{G}(q)(\hat{G}(v))^T.
\end{equation}

\textbf{Fusion search (ITV$_{fusion}$)}. As concept-based search and embedding search are complementary \cite{dual_task,tpami21-dual-encoding}, their search results are fused with a linear function, as follows:
\begin{equation}
\begin{aligned}
\label{eq:score_fusion}
Score(q,v)_{fusion} = (1-\theta)(Score(q,v)_{embedding}) + \theta(Score(q,v)_{concept}).
\end{aligned}
\end{equation}
where $\theta$ is a hyper-parameter to weigh the relative importance between the two search paradigms.

\section{Experiments}
\label{Sec:experiment}
We first evaluate our model on the MSR-VTT video captioning dataset \cite{msr-vtt} (Section \ref{Sec:t2vmsrvtt}) and then TRECVid video search datasets \cite{Trecvid2016,V3C1} (Section \ref{exp:AVS_TRECVid}). Next, the ablation studies are presented in Section \ref{sec:ab_study}. Finally, we exploit the most recent pre-trained features, i.e., CLIP \cite{CLIP} and BLIP \cite{BLIP2}, in our method and discuss their impact on search performance.

\subsection{Experimental settings}
 \label{Sec:experiment_setting}

\textbf{Concept bank}. The vocabularies are composed of words that appear more than five times in the training data of a video captioning dataset. Stop word removal is applied to exclude high-frequency words, while lemmatization is performed on verbs to derive their base forms. Following \cite{ACL:contradiction_in_text}, we use WordNet~\cite{wordnet} to automatically identify antonyms $T_i$ for each concept $i$ in the concept bank. For example, \textit{arise} and  \textit{stand} are detected as antonyms for \textit{sit}. Furthermore, each antonym's hyponymy is checked to identify the candidate pairs that are globally exclusive automatically. The pairs whose antonyms belong to either space (e.g., \textit{interior}, \textit{inside}) or time (e.g., \textit{daytime}, \textit{dawn}) relevant are retained for manual investigation. Finally, the concept pairs that are manually verified as not likely to co-exist in a video are identified as globally exclusive pairs, while the remaining pairs are declared as locally exclusive. In the experiments, we do not differentiate globally or locally exclusive concepts when applying Eq.~(\ref{eq:unlikelihood}). 

\textbf{Model training}. 
\label{Sec:model_training}
Each training video usually comes along with multiple captions. We pool all the captions of a video as the ground truth for interpretation training. Four types of deep features are extracted from ResNet-152 \cite{resnet}, ResNext-101 \cite{resnext}, swin-transformer \cite{SwinTrans} pre-trained on ImageNet-22K \cite{imagenet}, and SlowFast network \cite{slowfast} pre-trained on Kinetic-400 \cite{kinetics400}, respectively. The parameters of these networks are frozen during training. The model embeds the input dimensions 7,936 to $d=2,048$. The hyper-parameter $\lambda$ in Eq.~(\ref{eq:BCEAll}) is set to 0.2, as suggested in \cite{dual_task}. The model parameters are set to $\alpha=0.01$ in Eq.~(\ref{eq:concept_loss}), $c=0.2$ in Eq.~(\ref{eq:tripletloss}), and $\theta=0.5$ in Eq.~(\ref{eq:score_fusion}), unless otherwise stated. The learning rate is 0.0002, and the batch size is 128.

\subsection{Text-to-video search on the MSR-VTT dataset}
\label{Sec:t2vmsrvtt}

\begin{table}[]
\caption{Result comparison on the MSR-VTT dataset. Most performances are cited directly from the published results, except those marked with *, which are the rerun version. The unavailable results are  marked with /.} 
\label{exp:msrvtt}
\centering
\begin{tabular}{l|ccccc}
\toprule
              & \multicolumn{5}{c}{Text-to-Video Retrieval} \\ \hline
              & R@1    & R@5    & R@10   & Med r   & mAP    \\
              \hline
VSE++ \cite{vse}        & 8.7    & 24.3   & 34.1   & 28      & 16.9   \\
W2VV++ \cite{w2vvpp}       & 11.1   & 29.6   & 40.5   & 18      & 20.6   \\
Dual encoding \cite{dualconding} & 11.1   & 29.4   & 40.3   & 19      & 20.5   \\

HGR \cite{hgr}          & 9.2    & 26.2   & 36.5   & 24      & /     \\
CE  \cite{TPAMI2021_weight_metric}          & 10.9   & 30.4   & 42.3   & /      & /     \\
Hybrid space \cite{tpami21-dual-encoding} & 11.6   & 30.3   & 41.3   & 17      & 21.2   \\
Dual task* \cite{dual_task}       &12.2  &  31.7   &  42.9    & 16      &   22.3             \\
SEA* \cite{tmm2021-sea}          & 12.3    & 32.1     & 43.5     & 15     & 22.3     \\
LAFF* \cite{LAFF}          & 11.6    &  31.6    & 43.0      &  15      &   21.7    \\
RIVRL* \cite{RIVRL}         & 12.5    &  32.8    &   44.3    &   14     &  22.7      \\
 \hline
ITV         & 13.2   & 33.7   & 45.0   & 14      & 23.4 \\
\bottomrule
\end{tabular}
\end{table}

We conduct the experiment on MSR-VTT \cite{msr-vtt} using the official split of training, validation, and testing sets. A concept bank composed of 6,765 words with 902 exclusive pairs is formed for this experiment. Table \ref{exp:msrvtt} shows the performance of the proposed ITV model (i.e., ITV$_{fusion}$) in comparison to the existing works. Note that higher values of recall (i.e., R@{1,5,10}) and lower values of median rank (i.e., Med r) suggest better performance. It is worth noting that the results of SEA \cite{tmm2021-sea}, LAFF \cite{LAFF}, and RIVRL \cite{RIVRL} are obtained by rerunning the public codes released by the authors (marked with~*). For a fair comparison\footnote{The reported results by LAFF and RIVRL are (23.8, 49, 60.3, 6, 0.358) and (13.8, 35.0, 46.5, 13, 24.3) on (R@1, R@5, R@10, Med r, mAP) for use of (BoW, w2v, GRU, CLIP, BERT) and (BoW, w2v, GRU, BERT), respectively. In this paper, CLIP and BERT features are not used for LAFF and RIVRL for a fair comparison with other approaches.}, all the methods use the same textual encoder. Additionally, LAFF also employs the same visual features as the proposed ITV model. ITV outperforms all other approaches with a mean average precision (mAP) of 23.4. Randomization test \cite{randomization_test} shows that the differences in mAP are significant when comparing ITV with dual-task (i.e., the conference version), SEA, LAFF, and RIVRL at the level of p-value$\leq0.05$.

\begin{figure*}[t]
    \centering
    \includegraphics[width=1\linewidth]{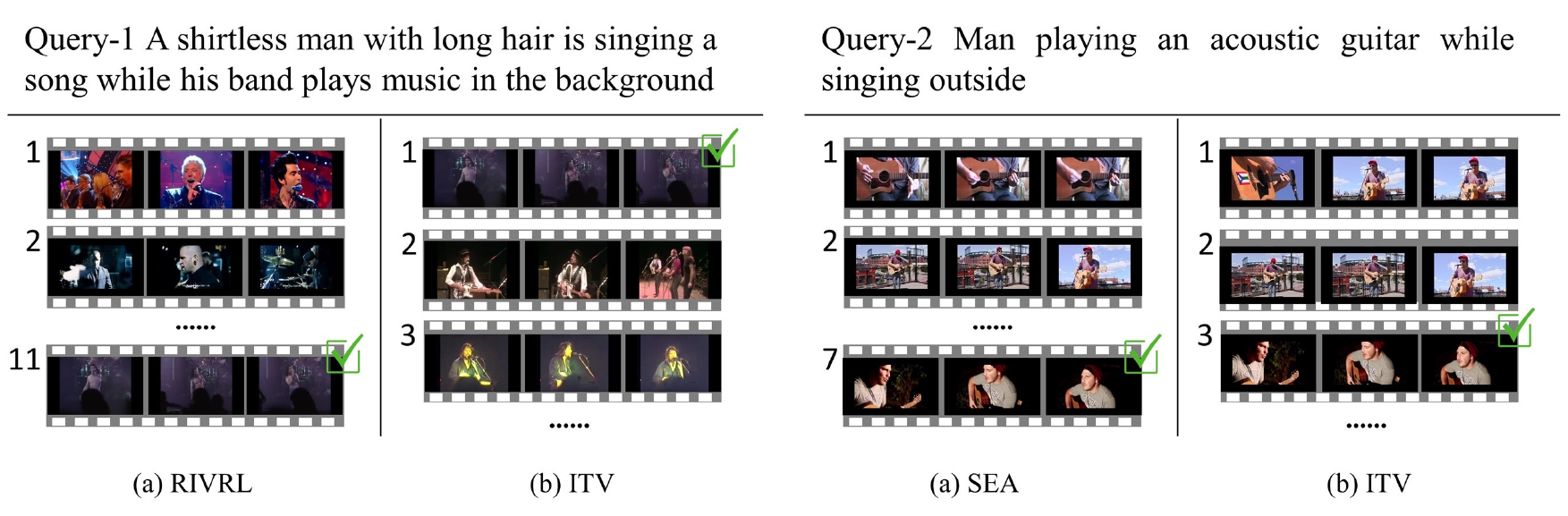}
    \caption{Two examples contrasting the search results of ITV with RIVRL and SEA. The ground-truth videos are marked with green ticks.}
    \label{fig:text2video_msrvtt}
\end{figure*}

Fig.~\ref{fig:text2video_msrvtt} illustrates two example queries that compare the performance of ITV with RIVRL and SEA. With unlikelihood training, in Query-1, ITV outperforms RIVRL by successfully ranking shots featuring a long-haired man higher, whereas RIVRL ranks shots with a short-haired man in the top two positions. In Query-2, ITV manages to rank the ground-truth shot at the third position by downgrading shots of playing guitar indoors. It is worth mentioning that even though the first two video shots are not ground truth, they match all the information required by the query. In contrast, the shots retrieved by SEA are mixed with indoor and outdoor scenes.

\subsection{Ad-hoc video search on TRECVid AVS datasets}
\label{exp:AVS_TRECVid}

As TRECVid AVS datasets do not come along with training data, the proposed model is trained on three video captioning datasets: TGIF \cite{tgif}, MSR-VTT \cite{msr-vtt}, and VATEX \cite{VATEX}. The resulting concept bank is composed of 14,717 words with 3,263 exclusive pairs. The VTT dataset \cite{Trecvid2016} released by TRECVid is used as a validation set in the experiments. The results are reported on the IACC.3 \cite{Trecvid2016}, V3C1 \cite{V3C1}, and V3C2 \cite{V3C1} of TRECVid AVS datasets, where a total of 190 queries are released for evaluation over a period of seven years (2016 to 2022). The video shots on these datasets are 335k (IACC.3), one million (V3C1), and 1.4 million (V3C2), respectively. Following the TRECVid standard \cite{Trecvid2016}, the mean extended inferred average precision (xinfAP) with a search length of 1,000 is used as the evaluation metric.


\begin{table*}[t]
\caption{Performance comparison across seven years of TRECVid AVS datasets. The number inside parentheses indicates the number of queries being evaluated in that year. The reproduced results are marked with *, while the unavailable results are indicated by /.} 
\label{tab:avs_comparison}
\centering
\begin{tabular}{l|ccc|ccc|c}
    \toprule
Datasets                     & \multicolumn{3}{c|}{IACC.3}               & \multicolumn{3}{c|}{V3C1} &V3C2    \\
\hline
Query sets                         & tv16 (30)         & tv17 (30)           & tv18  (30)          & tv19 (30)   &tv20 (20) &tv21 (20) &tv22 (30)      \\
                          \hline
\multicolumn{5}{l}{TRECVid top results:}                                                  \\
Rank 1     & 0.054  & 0.206  & 0.121 & 0.163   &\textbf{0.359}   &\textbf{0.355} \cite{vireo_2021} & \textbf{0.282} \cite{Waseda2022} \\
Rank 2     & 0.051   & 0.159   & 0.087  & 0.160    &0.269   & 0.349     & 0.262 \\
Rank 3     & 0.040   & 0.120   & 0.082       & 0.123     &0.229     &0.343 & 0.210 \\
\hline
\multicolumn{5}{l}{Concept-based search:} \\
QKR \cite{QKR:AVS}  & 0.064          & /              & /              & /      & /    & /    & /   \\
ExclusiveConcept \cite{Kindai_kobe2018}  & /          & /              & 0.072              & /  & /   & /   & /   \\
ConBank (auto)  & /  & 0.159 \cite{Waseda_Meisei2017}  & 0.060 \cite{Waseda2018}        & /    & /     & /     & /   \\
ConBank (manual)   & 0.177 \cite{Waseda2016}   & 0.216 \cite{Waseda_Meisei2017}  & 0.106 \cite{Waseda2018}        & 0.114  \cite{WasedaMeiseiSoftbank2019}  & 0.183 \cite{wasedaAVS2020}    & /  & /    \\

Dual-task$_{concept}$ \cite{dual_task}              & 0.138          & 0.156           & 0.081        &0.110    & 0.192   & 0.167  & / \\ 
Hybrid space$_{concept}$* \cite{tpami21-dual-encoding}              & 0.119          & 0.188           & 0.084         & 0.120    & 0.210  & 0.189 &0.096  \\ 
RIVRL$_{concept}$* \cite{RIVRL}  &  0.119          &      0.188          &    0.092           & 0.126      & 0.225   & 0.182 &  0.104\\
\hline
\multicolumn{5}{l}{Embedding search:} 
\\
VideoStory \cite{videostory} & 0.087          & 0.150           & /              & /          & /  & /  & /  \\
VSE++ \cite{vse}           & 0.135          & 0.163          & 0.106         & /    & /     & /     & /     \\
W2VV \cite{w2vv}                      & 0.149          & 0.198          & 0.103          & /  & /   & /   \\  
W2VV++ \cite{w2vvpp}       & 0.150         & 0.207           & 0.099         & 0.146     & 0.199      & /  & /  \\

Dual coding \cite{dualconding}              & 0.160          & 0.232           & 0.120         & 0.163  & 0.208    & /    & / \\ 

HGR \cite{hgr}             &/         & /           & /       & 0.142   &0.301     & /  & /   \\ 
Dual-task$_{embedding}$ \cite{dual_task}              & 0.161          & 0.239           & 0.113         &0.170    & 0.202   & 0.167  & / \\ 
Hybrid space$_{embedding}$* \cite{tpami21-dual-encoding}              & 0.144         & 0.228           & 0.104         & 0.166    & 0.243   & 0.218  & 0.118\\ 
SEA* \cite{tmm2021-sea}             & 0.179         & 0.289           & 0.136       & 0.191     & 0.327  & 0.292 & 0.120 \\ 
RIVRL$_{embedding}$* \cite{RIVRL}  &   0.140         &      0.216          &  0.110             &  0.154     &  0.251  &  0.197  &0.126     \\
LAFF* \cite{LAFF}             &     0.173   &   0.263         &  0.138      &  0.175   &  0.270  & 0.255 &0.102  \\ 
\hline

\multicolumn{5}{l}{Fusion search:} \\
Dual-task$_{fusion}$ \cite{dual_task}              & 0.184          & 0.252           & 0.120        &0.189    & 0.229    & 0.193 & /\\ 
SEA-hybrid \cite{tmm2021-sea}             & 0.166         & 0.235           & 0.126       & 0.172   & 0.196 & /  & /  \\ 
SEA+dual encoding \cite{tmm2021-sea}             & 0.173         & 0.250           & 0.128       & 0.171   & 0.209   & /  & /  \\ 
Hybrid space$_{fusion}$* \cite{tpami21-dual-encoding}              & 0.144          & 0.228           & 0.105         & 0.165    & 0.246  & 0.218 & 0.120 \\
RIVRL$_{fusion}$* \cite{RIVRL}  &  0.137          &       0.212         &      0.108         &  0.145     &  0.248  & 0.195 & 0.120 \\
\hline

\multicolumn{5}{l}{The proposed models:}  
                               \\
ITV$_{concept}$            &  0.184     & 0.230   & 0.135      & 0.166          &  0.292    & 0.246  & 0.115 \\
ITV$_{embedding}$           &  0.187    &   0.279  &  0.140   &  0.201          & 0.307    &   0.294  &0.135 \\
ITV$_{fusion}$             &  \textbf{0.211}     &  \textbf{0.292}    &   \textbf{0.170}  &   \textbf{0.277}          &0.345  & 0.318 & 0.150 \\
\hline

\bottomrule
\end{tabular}
\end{table*}

Table \ref{tab:avs_comparison} lists the ITV results in comparison to the existing works and the best three reported results by TRECVid in each year of evaluation. Among the compared works, W2VV++ \cite{w2vvpp} is the first embedding search that shows significant improvement over the concept-based search. 
SEA \cite{tmm2021-sea}, hybrid space \cite{tpami21-dual-encoding}, LAFF \cite{LAFF}, and RIVRL \cite{RIVRL} models are currently the state-of-the-art techniques for AVS. The concept-based model QKR \cite{QKR:AVS} performs a series of NLP steps to identify concepts in word, phrase, and sentence levels for concept selection. ExclusiveConcept \cite{Kindai_kobe2018} conducts manual concept selection and uses four exclusive concept pairs to filter out false positives. A large concept bank (ConBank) consisting of over 55 thousand words is used by \cite{Waseda2016,Waseda_Meisei2017,Waseda2018,WasedaMeiseiSoftbank2019,wasedaAVS2020} for automatic and manual selections of concepts. The hybrid approaches based on late fusion are compared, including Dual-task$_{fusion}$, SEA-hybrid, SEA+dual encoding, hybrid space$_{fusion}$, and RIVRL$_{fusion}$. For those models where the codes are publicly available, we rerun the models using the same setting as ours for fair comparisons (e.g., the same training sets and textual/visual features) and report the results across seven years of query sets (tv16 to tv22).

As shown in Table \ref{tab:avs_comparison}, ITV$_{concept}$ outperforms the traditional concept-based approaches, including QKR, ExclusiveConcept, and ConBank with automatic concept selection, by a large margin. The result is even better than ConBank, which manually picks concepts from a bank of size four times larger than ITV. Compared to the most recent concept-based approaches, ITV$_{concept}$ shows better xinfAP in 135 out of 190 queries than hybrid space$_{concept}$ and in 102 queries than RIVRL$_{concept}$. A similar performance trend is also observed when comparing ITV$_{embedding}$ to the existing embedded-based search approaches. ITV$_{embedding}$ outperforms hybrid space$_{embedding}$, SEA, RIVRL$_{embedding}$, and LAFF. Particularly, ITV$_{embedding}$ shows better performance in 149 out of 190 queries than RIVRL$_{embedding}$ and in 118 queries than LAFF. Indeed, using either concept or embedding search by ITV has already attained results better than most of the fusion search methods such as SEA+dual encoding, hybrid space$_{fusion}$, and RIVRL$_{fusion}$. When combining concept and embedding searches, ITV$_{fusion}$ achieves either the best or very competitive performances among the published results across seven years of query sets evaluated in TRECVid. Using the randomization test \cite{randomization_test}, the results of ITV are shown to be significantly different from other counterparts in concept-based, embedding, and fusion search, respectively, at the level of p-value $\leq$ 0.01. Note that the top-rank official results published by TRECVid are all based on the ensemble of different models, which are expected to exhibit stronger performances. For example, the top-performing result on tv21 \cite{vireo_2021} is based on the fusion of ITV$_{fusion}$, dual coding, and dual-task with phrase decoding, and the top-1 result on tv22 \cite{Waseda2022} is based on the fusion of VSE++ \cite{vse}, GSMN \cite{GSMN}, CLIP \cite{CLIP}, SLIP\cite{SLIP} and diffusion models. 

To show the complementary power of ITV with other methods, we perform a late fusion of ITV and \cite{Waseda2022} (the top-1 performer of TRECVid 2022) by an average linear combination of their search rank lists. The fusion results in around a 10\% performance boost over \cite{Waseda2022}, with 23 out of 30 queries attaining higher xinfAP. Both rank lists share 88\% of true positive shots, and the improvement is mainly due to the elevation of some positive shots to higher ranks. For example, among the queries with improved performance, the average number of true positives with higher ranking positions in the search depth of 10 is 4.13. With late fusion, ITV is also able to retrieve more true positive shots (ranging from 1 to 90) originally outside the search depth of 1,000 for all queries.

\begin{figure*}[t]  
\centering  
  \centering  
  \includegraphics[width=1\linewidth]{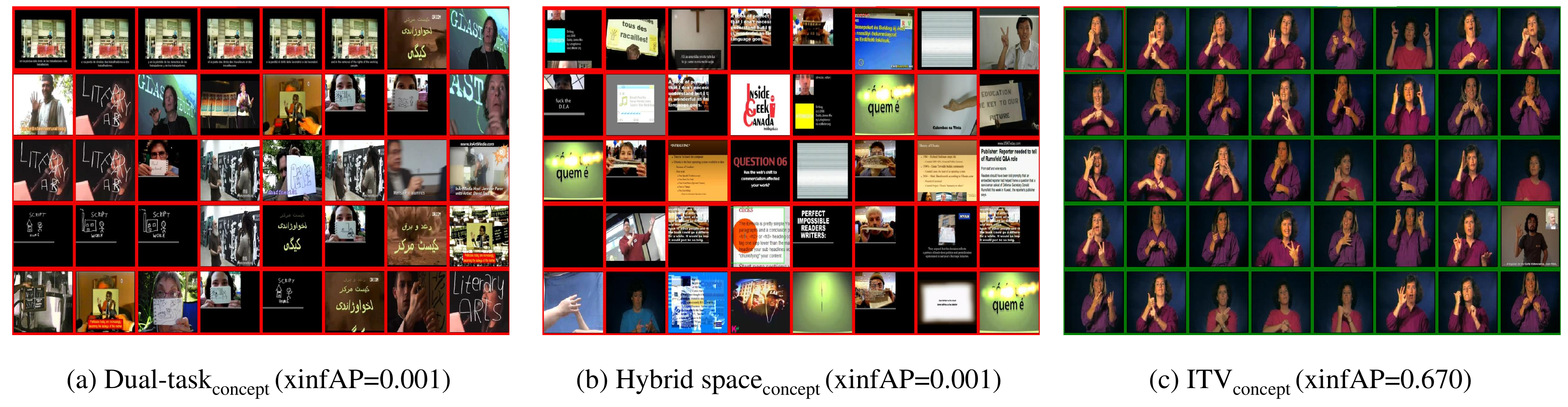}  
    \caption{Comparison with concept-based search methods (dual-task$_{concept}$ and hybrid space$_{concept}$) on the query-543 \textit{Find shots of a person communicating using sign language}.}   
    \label{fig:sign_langage}
\end{figure*}

\begin{figure*}[t]  
\centering  
  \centering  
  \includegraphics[width=1\linewidth]{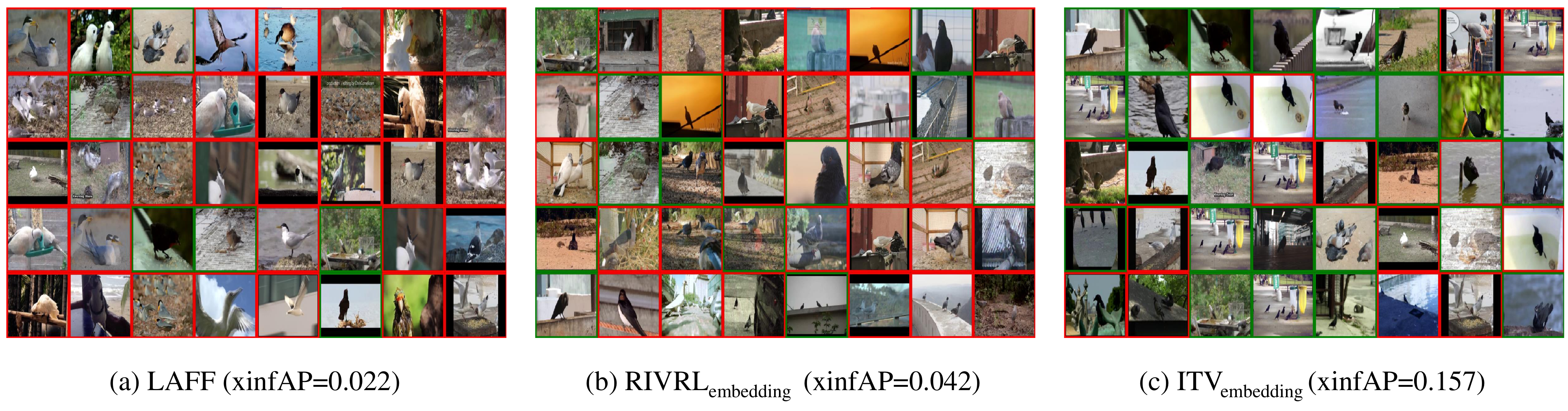}  
    \caption{Comparison with embedding search methods (LAFF and RIVRL$_{embedding}$) on the query-717 \textit{Find shots of a black bird seen on a dry area sitting, walking, or eating}.}   
    \label{fig:black_bird}
\end{figure*}

\begin{figure*}[t]  
\centering  
  \centering  
  \includegraphics[width=0.99\linewidth]{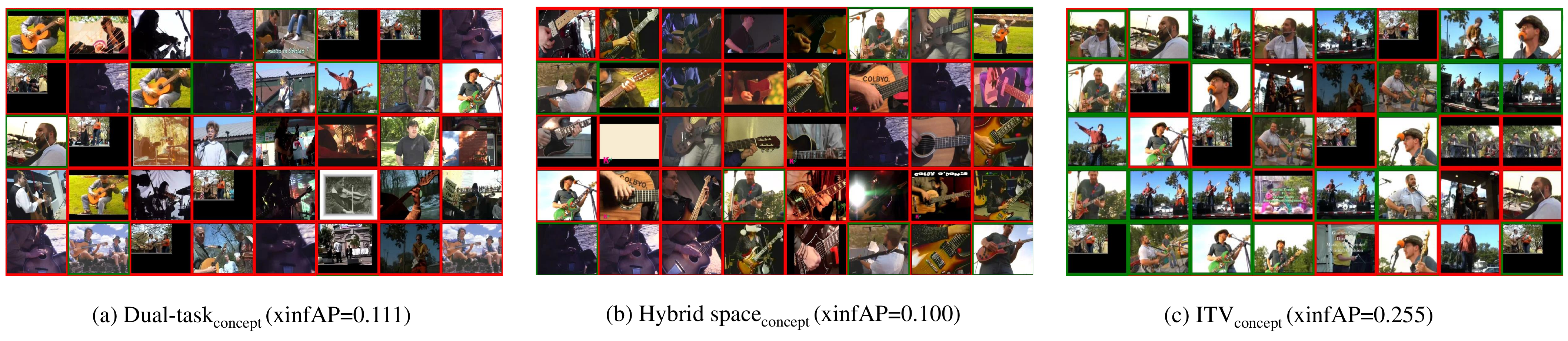}  
    \caption{Comparison with likelihood search methods (dual-task$_{concept}$ and hybrid space$_{concept}$) on the query-501 \textit{Find shots of a person playing guitar outdoors}.}   
    \label{fig:guitar}
\end{figure*}

Figures \ref{fig:sign_langage}-\ref{fig:guitar} show examples verifying the strength of the ITV model. In Fig.~\ref{fig:sign_langage}, the query term {\em sign language} cannot be properly expressed by other concept-based search methods, resulting in video shots showing hand gestures or text signs being retrieved. In contrast, ITV$_{concept}$ interprets the query with a concept list of [\textit{person, talk, use, language, sign, signs, communicate, communication, translate, interpret, deaf, interpreter, ASL}], which contains novel terms such as {\em deaf} and {\em ASL}, and successfully retrieves true positives, even including those that are not labeled in the ground-truth results. The unlikelihood model also benefits embedding search considerably, as shown in Fig.~\ref{fig:black_bird}, for the query to search for a black bird. While most approaches manage to retrieve shots with a white bird, ITV$_{embedding}$ is the method that ranks the most number of the shots with a black bird to the top of the list. We attribute the high performance due to faithful interpretation of the query embedding by ITV to capture the information need, where the top-10 decoded concepts are [\textit{bird, birds, black, eat, walk, food, area, dry, peck, sit}], and simultaneously assign the exclusive concepts \textit{[white, wet]} with low probability scores. The unlikelihood model shows advantages for queries specifying constraints such as indoors or outdoors, as shown in Fig.~\ref{fig:guitar}. ITV$_{concept}$ interprets the query with the concept {\em outdoors} in high probability while lowering the probability of the concept {\em indoors}. Therefore, the video shots with a man playing guitar outdoors are ranked higher than the existing AVS methods that also decode concepts from embedding (i.e., dual-task \cite{dual_task} and hybrid space \cite{tpami21-dual-encoding}).

\subsection{Ablation studies}
\label{sec:ab_study}
In the following, we verify the contributions of various components newly introduced to ITV and make comparisons with our conference version \cite{dual_task}. In all the experiments, the ITV models are trained following the same experimental setting and training datasets as the previous section (Section \ref{exp:AVS_TRECVid}).

\subsubsection{Elimination of Exclusive Concepts}
\label{Sec:elimination_concept}

We first assess the effect of unlikelihood training in eliminating the exclusive concepts from interpreting an embedding. For example, when only the action {\em sit} is observed in a video shot, we study whether the concept {\em stand} will be suppressed from interpretation by setting the threshold of not decoding a concept $i$ with $\hat{g}_i< 0.5$. We conduct experiments on MSVD \cite{msvd} and MSCOCO \cite{MSCOCO} for video and query embedding interpretation, respectively.

MSVD \cite{msvd} is composed of 1,970 videos, with each video being associated with around 40 captions. We report the average success rate of concept suppression for all the 1,153 exclusive concept pairs identified for this dataset. On average, each pair has 100 videos where their captions contain one of the words in the pair. We compare the result to the model where only likelihood loss is considered (i.e., \cite{dual_task}). The success rate of eliminating exclusive concepts is as high as 0.87, in contrast to the rate of 0.70 attained by ITV without unlikelihood training. The performance gain is considered significant. For example, in the absence of unlikelihood training, the concept {\em woman} is decoded for 756 videos with only man visible. The false decoding is reduced significantly to 271 videos when ITV is trained with both likelihood and unlikelihood losses. Note that, for locally exclusive pairs, it is possible that both concepts in a pair are mentioned in the captions of a video. To assess the chance that those locally exclusive concepts are erroneously removed, we also measure the missing rate for 107 exclusive pairs. On average, each pair involves 28 testing videos. The missing rates are 0.29 and 0.19 with and without unlikelihood training, respectively. A higher rate is noted for unlikelihood training for video examples where the subject of interest is partially visible. For example, {\em man} is mistakenly suppressed from a music video because only posture is visible but not face. There are a few extreme cases, such as {\em A 12-year-old young boy}, where \textit{old} is not decoded by unlikelihood training.  

MSCOCO \cite{MSCOCO} is composed of 616,838 captions, and a total of 1,498 exclusive pairs are identified for evaluating query interpretation. On average, each pair involves 796 captions. The success rate of suppressing exclusive concepts is 0.94, with a missing rate of 0.036. In the absence of unlikelihood training (i.e., \cite{dual_task}), the success rate drops to 0.73, with a lower missing rate of 0.021. Similar to video embedding interpretation, the slightly higher missing rate in unlikelihood training is mostly due to extreme examples. For example, the concept {\em black} is not interpreted for the caption {\em an open white toilet against a wall of white tiles and a black and white floor}, which involves logical reasoning.

\subsubsection{Query Interpretation for Concept-based Search}
\label{Sec:query_interpretation}

Next, we evaluate the extent to which relevant concepts can be decoded to subtly explain query embedding and provide insights on how the interpretation can lead to stronger search performance. We compare ITV with and without unlikelihood loss (UL) training to three concept selection methods. The methods are BERT \cite{sentence-bert}, word-to-vec (w2v), and its variant i-w2v, which is proposed in \cite{Boer:Semantic} to address the problem of query drift. These methods aim to select a small set of concepts similar to query tokens for similarity measure as Eq.~(\ref{eq:score_concept}). The methods BERT and w2v exhaustively measure the BERT/word-to-vec similarity between the query sentence and each concept and then select the top-k most similar concepts. Here, $k$ is set to be the number of query tokens. Instead of fixing the value of $k$, the method i-w2v adaptively determines $k$ depending on query content. By treating a query as a sentence, i-w2v incrementally picks one concept that is most similar to the query sentence. The process stops when the inclusion of the next concept will downgrade the similarity of concepts being selected so far to the query. For a fair comparison, the videos are indexed with concepts decoded by ITV with UL. In other words, the only variable in this experiment is the use of different models to select concepts from user queries for video search. 

\begin{table}[t]
\caption{Performance comparison across different concept selections on two AVS benchmark datasets.} 
\label{exp:comparison_concept_search}
\centering
\begin{tabular}{c|ccc|ccc|c}

    \toprule
                                                 & \multicolumn{3}{c|}{IACC.3 dataset} & \multicolumn{3}{c|}{V3C1 dataset}\\\hline
Concept selection                       & tv16       & tv17      & tv18      & tv19            & tv20   & tv21      & average    \\\hline
w2v         & 0.132      & 0.163     & 0.089     & 0.103          & 0.171    &0.178    &0.139      \\
i-w2v \cite{Boer:Semantic}  &0.143      & 0.167   & 0.084     &  0.099          &0.172    &0.224   & 0.148    \\
BERT \cite{sentence-bert}        & 0.153      & 0.138     & 0.097     & 0.088          &0.225 &0.238 &0.156    \\ 

\hline
ITV w/o UL  \cite{dual_task}         & 0.159      &0.193     &0.101     &0.150          & 0.218 & 0.219  &0.173\\  
ITV w/ UL          & \textbf{0.184}      & \textbf{0.230}     & \textbf{0.135}     & \textbf{0.166}           &\textbf{0.292}  & \textbf{0.246}     &\textbf{0.209}     \\ 
\bottomrule  
\end{tabular}
\end{table}

\begin{figure*}[t]  
\centering  
\includegraphics[width=1\linewidth]{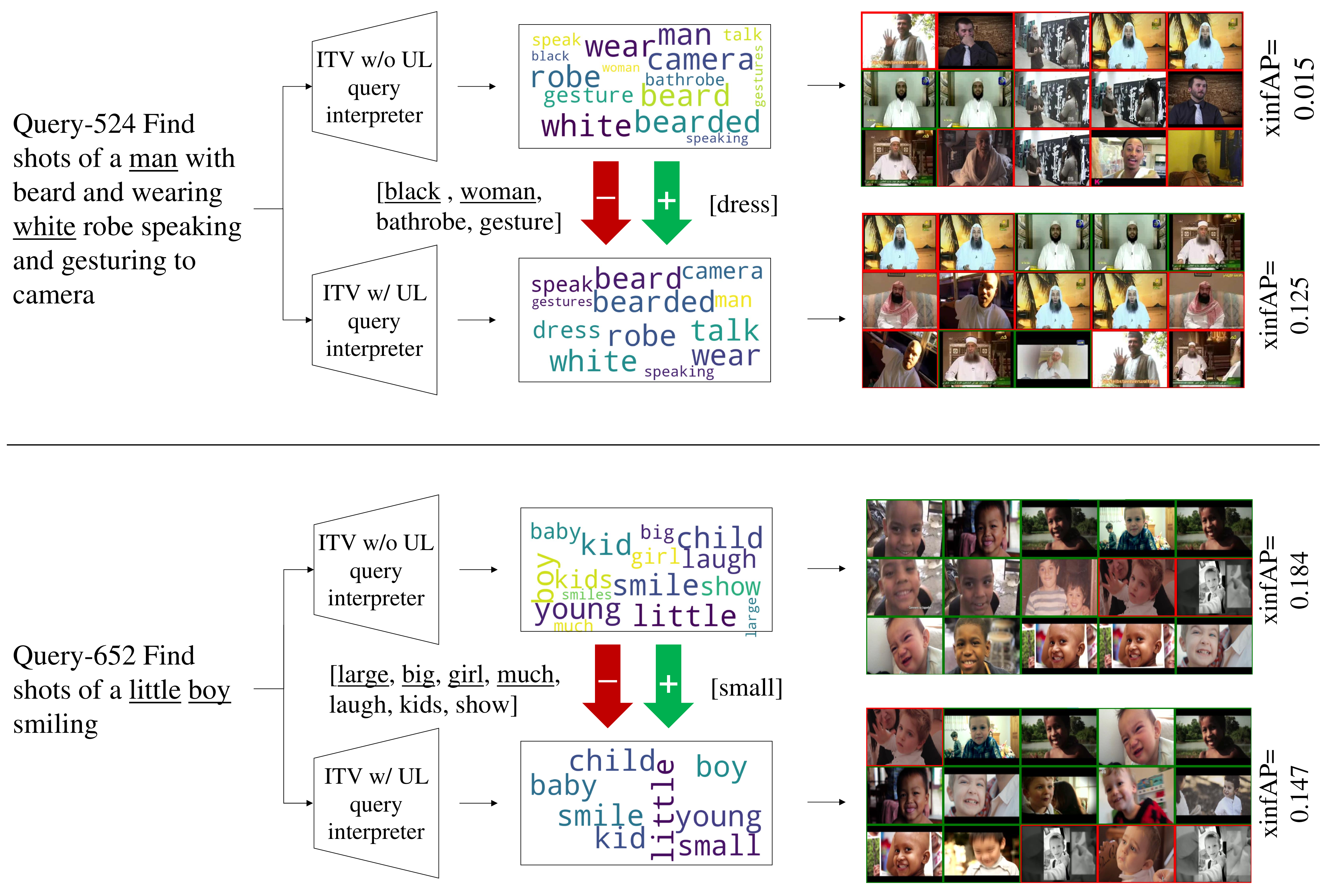}
\caption{Examples illustrating the quality of query embedding interpretation. The red (green) arrow shows the concepts being excluded (included) with unlikelihood training. Exclusive concepts are underlined for ease of reading.} 
\label{fig:nonUL_VS_UL}
\end{figure*}

As shown in Table~\ref{exp:comparison_concept_search}, ITV consistently outperforms other approaches. With UL, the performance of ITV generally benefits from a higher quality of embedding interpretation. The number of decoded concepts is 2.57 less than ITV without UL \cite{dual_task}. Fig.~\ref{fig:nonUL_VS_UL} shows two examples contrasting the performances. In query-524, the concepts {\em black} and {\em woman} are successfully removed by UL, resulting in a significant boost by downgrading the ranks of videos with persons wearing a robe not in white color. Nevertheless, in query-652, despite successfully excluding concepts like {\em large} and {\em girl}, the concept {\em laugh}, which is critical in describing {\em a boy smiling}, is also erroneously removed. Consequently, the search performance of ITV with UL is hurt for this query. It is worth noting that in these two examples, some coherent concepts are also added by likelihood loss as pull-push perturbations. Overall, ITV with UL shows better xinfAP in 81 out of 115 queries that involve exclusive concepts.

\begin{figure*}[t]  
\centering  
  \centering  
  \includegraphics[width=0.8\linewidth]{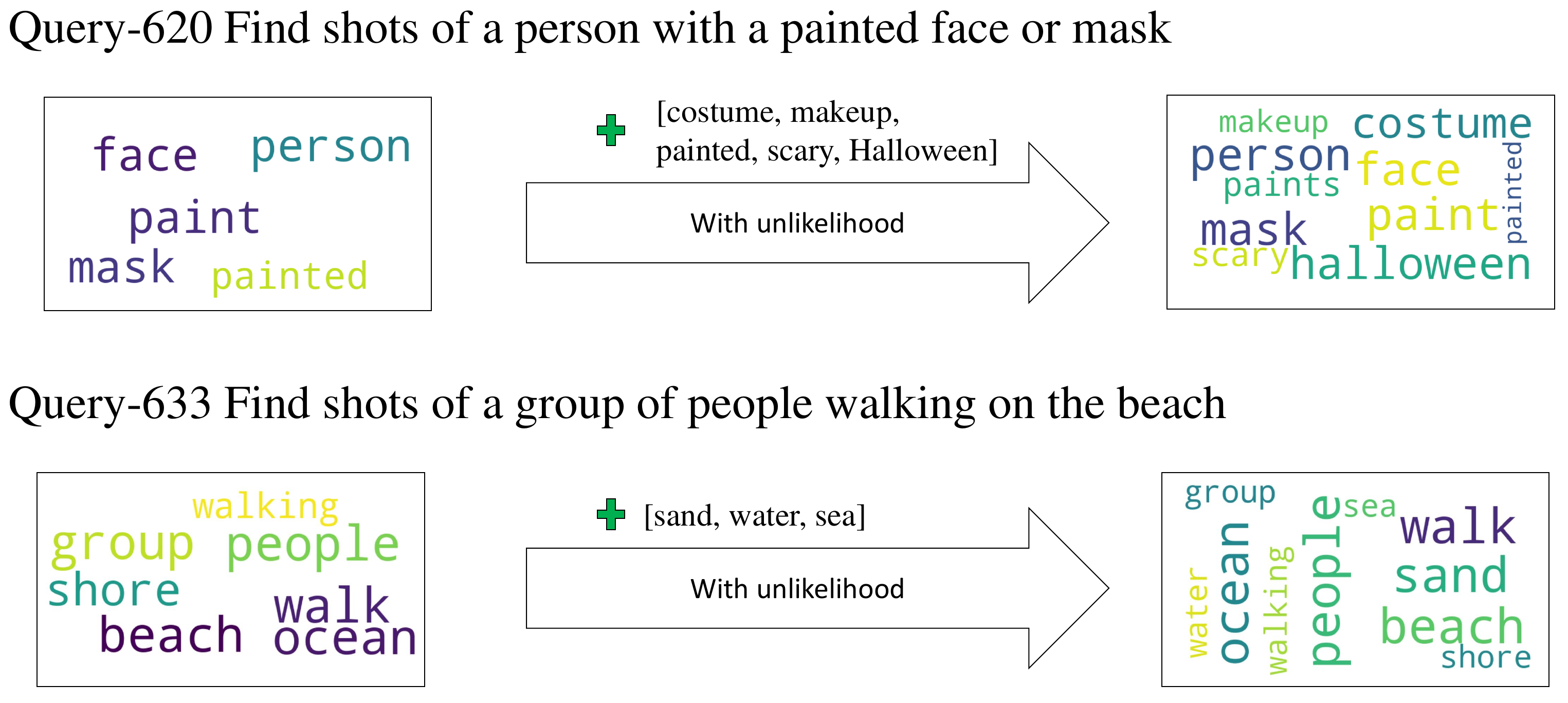}  
    \caption{Two examples comparing the query embedding interpretations generated by the models without (left) and with (right) unlikelihood training.}   
    \label{fig:query_interpretation_ULvsnonUL}
\end{figure*} 

\begin{figure*}[t]  
\centering  
\includegraphics[width=1\linewidth]{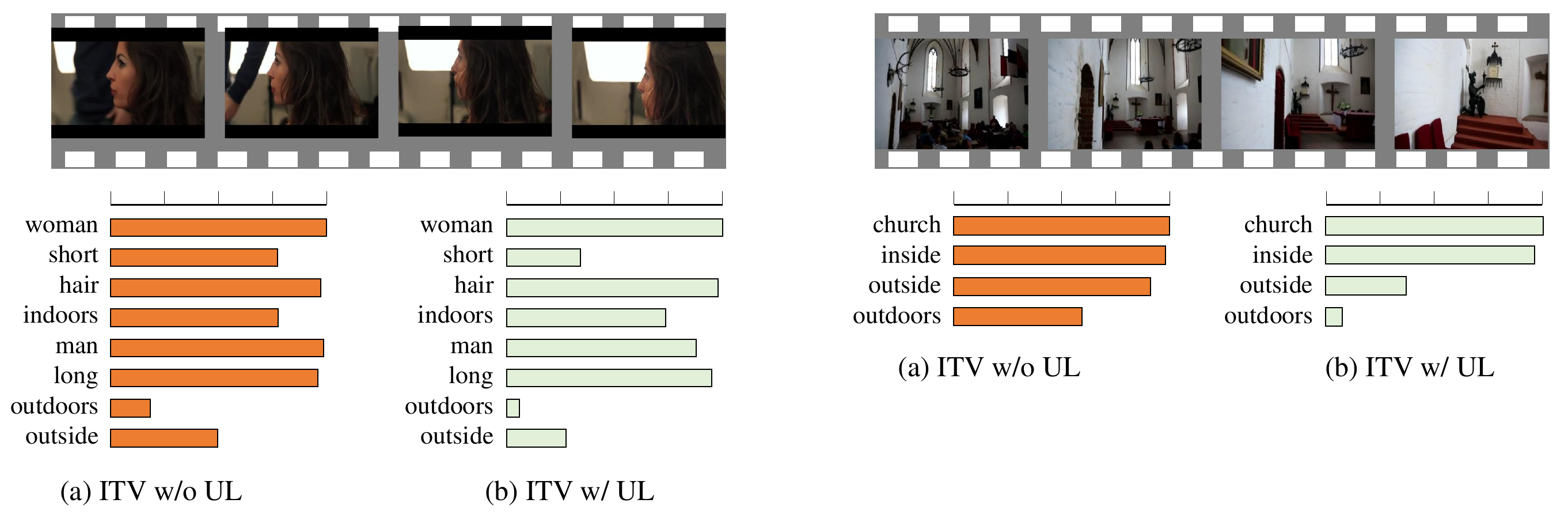}
\caption{Two video examples showing the probability distribution of concepts being decoded from their video embeddings: a long-hair woman indoors (left) and a church from the inside (right).} 
\label{fig:visual_interpretation}
\end{figure*}

Unlikelihood training also improves the interpretability of query embedding and video embedding, as shown in Fig.~\ref{fig:query_interpretation_ULvsnonUL} and Fig.~\ref{fig:visual_interpretation}, respectively. As shown in Fig.~\ref{fig:query_interpretation_ULvsnonUL}, some novel terms are introduced by unlikelihood training, such as [\textit{costume, makeup, painted, scary, Halloween}] for the query finding a person with a painted face or mask, and [\textit{sand, water, sea}] for the query finding a group of people walking on the beach. These additional words supplement the original queries with contextually relevant information. Fig.~\ref{fig:visual_interpretation} compares the video embedding interpretations generated by ITV with and without unlikelihood loss. The first example shows a video clip with a long-haired woman indoors. With UL, the probabilities of concepts such as \textit{short}, \textit{outdoor}, and \textit{outside} become relatively lower. The second video clip shows a church from the inside. UL is able to decode the concepts \textit{church} and \textit{inside} with high probability while keeping the probabilities of \textit{outdoors} and \textit{outside} low.

\subsubsection{Impact of the Proposed Unlikelihood Loss on Search} 
\label{Sec:unlikelihood_loss_on_search}

\begin{table}[t]
    \caption{Performance difference contrasting ITV with and without unlikelihood training. The number in bold shows a relatively better performance. (UL: Unlikelohood training with Eq. (4); $\widehat{UL}$: Unlikelihood training with Eq. (3); -UL: without unlikelihood training)}
    \centering
\begin{tabular}{c|ccc|ccc|ccc}
\toprule
        & \multicolumn{3}{c|}{Concept-based search} & \multicolumn{3}{c|}{Embedding search} & \multicolumn{3}{c}{Fusion search} \\ \hline
        & -UL   & $\widehat{UL}$   &  UL     & -UL    & $\widehat{UL}$   &  UL      & -UL    & $\widehat{UL}$   &  UL      \\\hline
tv16     
& 0.171   &0.180       & \textbf{0.184}        
& 0.186   &0.178         & \textbf{0.187}          
& 0.200   &0.203      & \textbf{0.211}           
\\
tv17     
& 0.208    &0.203        & \textbf{0.230}           
& 0.268    &0.277         & \textbf{0.279}            
& 0.286    &0.281       & \textbf{0.292}         
\\
tv18    
& 0.114     &0.123       & \textbf{0.135}          
& 0.129     &0.138        & \textbf{0.140}            
& 0.145     &0.158      & \textbf{0.170}          
\\
tv19    
& \textbf{0.167}    &0.152             & 0.166         
& 0.195             &\textbf{0.208}    & 0.201         
& 0.210             &0.222             & \textbf{0.227}           
\\
tv20    
& 0.266            &0.256  & \textbf{0.292}           
& \textbf{0.313}   &0.308            & 0.307           
& 0.334            &0.327  & \textbf{0.345}          
\\
tv21     
& 0.247            &\textbf{0.249}  & 0.246             
& \textbf{0.294}   &0.272           & \textbf{0.294}           
& \textbf{0.318}   &0.304           & \textbf{0.318}         
\\ \hline
average  
& 0.195           &0.194  & \textbf{0.209}            
& 0.231           &0.230  & \textbf{0.235}            
& 0.249           &0.249  & \textbf{0.261}     \\  

 \bottomrule  
\end{tabular}
\label{Tab:VT_UL_VT}
\end{table}

We investigate the effect of the proposed unlikelihood loss (UL) for concept-based, embedding, and fusion searches in comparison to the conference version \cite{dual_task} (-UL) on TRECVid AVS datasets. The results are listed in Table \ref{Tab:VT_UL_VT}. The performance gains due to UL are apparent across different search modes in most years of the query sets. Among the 115 queries involving at least one exclusive concept, the average gains are +587\% and +17\% for concept-based and embedding searches, respectively. For the remaining queries without exclusive concept, ITV with the proposed UL also receives gains of +33\% and +12\%. Overall, fusion search attains the best performance consistently across all the query sets. Randomization test \cite{randomization_test} shows that the differences between the UL and without UL in all three search modes are significant at the level of p-value $\leq 0.05$. To show the importance of considering context when applying penalty to an exclusive concept, we also compare the performance of unlikelihood training using Eq. (\ref{eq:oriUL}) and Eq. (\ref{eq:unlikelihood}), denoted as $\widehat{UL}$ and UL in Table \ref{Tab:VT_UL_VT}, respectively. The result shows the significance of adding the term (1-$g_t$) to enable selective matching of exclusive concepts in Eq. (\ref{eq:unlikelihood}). Without the term (1-$g_t$), the unlikelihood objective does not contribute (and even slightly degrades the performance of likelihood training sometimes) in most of the TRECVid AVS query sets.

The success of the proposed UL can be attributed to the higher quality in interpreting query and video embeddings. Consistent improvements are noted for both concept and embedding searches in 84 out of 160 queries in the comparison of ITV with and without UL. For example, in query-634 \textit{Find shots of a woman and a little boy both visible during daytime}, concept (embedding) based search attains a performance boost of 686\% (167\%) due to the introduction of new query terms ({\em toddler} and {\em mom}) and the removal of exclusive terms ({\em man}, {\em girl}, {\em daughter}, {\em big}, {\em dark}, {\em large}). Nevertheless, the degree of improvement varies across different search modes. For example, UL introduces a new term {\em computer} to query-570 {\em Find shots of a projection screen}. The performance of concept-based search is improved from 0.048 to 0.212 while embedding search only attains a slight improvement from 0.001 to 0.059.

However, both concept-based and embedding searches generally suffer when UL removes exclusive concepts that provide useful context for search. For example, the term {\em dark}, as an antonym of {\em light}, is suppressed by UL for query-529 {\em Find shots of a person lighting a candle}. For this particular query, the terms {\em dark} and {\em lighting} are helpful in searching for a moment when the unlit scenes are lighted up. ITV without UL is able to perform considerably better in this query for being able to introduce new query terms {\em dark} and {\em cake}. Overall, concept-based search suffers more than embedding search when the terms introduced by likelihood training are erroneously suppressed by UL. For example, in query-647 \textit{Find shots of people or cars moving on a dirt road}, even though the terms {\em dusty} and {\em terrain} are removed by UL, embedding search is still able to demonstrate competitive performance as ITV without UL. Concept-based search, however, degrades xinfAP by around 34\%.

We also investigate a different setting by mandatorily penalizing the globally exclusive concepts, or more specifically, by setting (1-$g_t$)=1 for all globally exclusive concepts when applying Eq. (\ref{eq:unlikelihood}). There is insignificant performance difference between this and the default setting (Section \ref{Sec:experiment_setting}). This is mainly due to the small number of globally exclusive concept pairs (i.e., 50 out of 3,263 exclusive pairs) being identified, which does not impact the performance positively or negatively.

\subsubsection{Sensitivity of $\lambda$.} 
\label{Sec:sensitity_of_lamda}
\begin{figure}[t]  
\centering  
\includegraphics[width=0.72\linewidth]{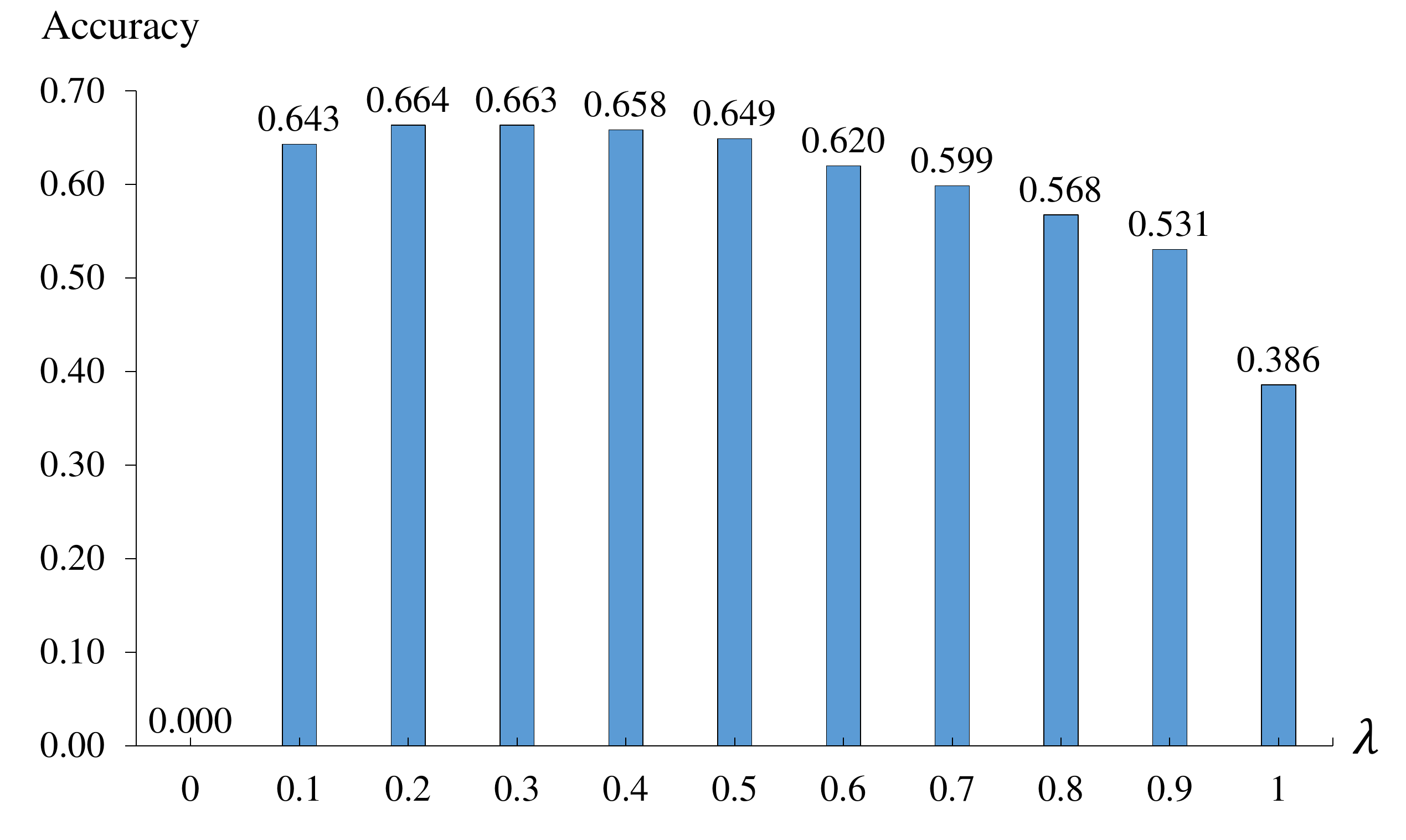}
\caption{Sensitivity of the hyper-parameter $\lambda$ on multi-class concept classification.} 
\label{fig:lambda_parameter}
\end{figure}

The hyper-parameter $\lambda$ in the likelihood training (i.e., Eq. (\ref{eq:BCEAll})) is to control the number of unmentioned concepts by the ground-truth labels to be decoded for an embedding. Specifically, a large value of $\lambda$ encourages more unmentioned concepts to be decoded or vice versa. In principle, $\lambda$ should be smaller than 0.5 to ensure only a small number of unmentioned but relevant concepts are activated for an embedding and enforce that all mentioned concepts can be decoded. To verify this, we compute the multi-class concept classification accuracy on both video and text embeddings using the MSVD dataset \cite{msvd} by varying $\lambda=\{$0, 0.1, 0.2, 0.3, 0.4, 0.5, 0.6, 0.7, 0.8, 0.9, 1.0$\}$ in Eq.(\ref{eq:BCEAll}). In total, MSVD has 1970 videos and 80, 837 captions. Fig. \ref{fig:lambda_parameter} compares the mean average accuracy of true positives at top-10 decoded concepts for both video and text embeddings. The result shows that $\lambda=0.2$ or $\lambda=0.3$ obtains the highest accuracy on decoding correct concepts in the top 10. For example, when $\lambda=0.2$, there are at least six positive concepts in the top 10, and the remaining four unmentioned concepts are also more likely to be positive than negative. For example, the concepts \textit{airplane} and \textit{flight} are not mentioned instead of irrelevant for a caption \textit{a plane is flying in the sky}. When $\lambda=0.0$, the performance is very low as no positive concepts are encouraged to be decoded in Eq. (\ref{eq:BCEAll}). In this paper, we use $\lambda=0.2$ as the default setting. 

\subsubsection{Sensitivity of $\alpha$.} 
\label{Sec:sensitity_of_alpha}

\begin{figure}[t]  
\centering  
\includegraphics[width=0.75\linewidth]{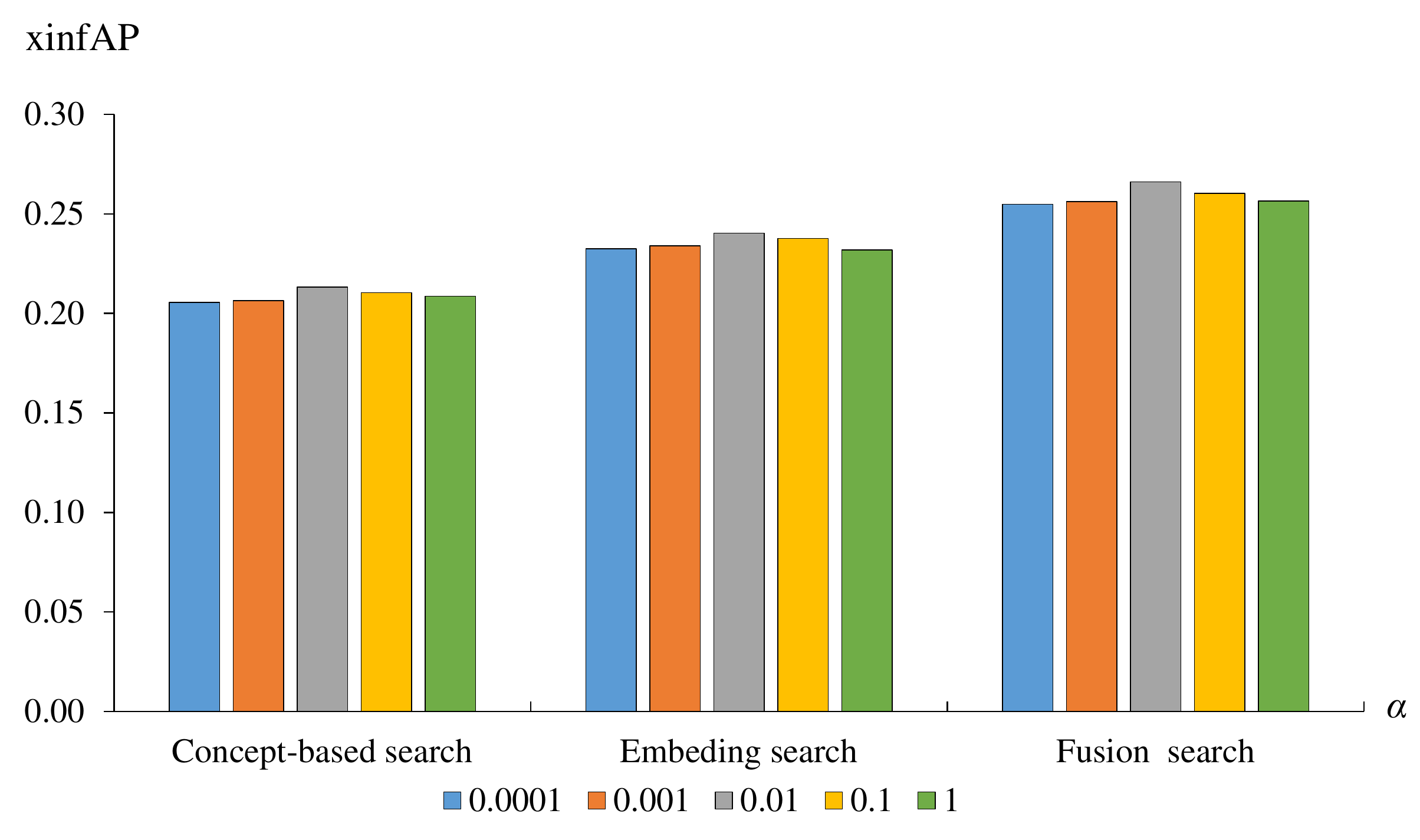}
\caption{Sensitivity of the hyper-parameter $\alpha$} 
\label{fig:alpha_parameter}
\end{figure}

The hyper-parameter $\alpha$ in Eq.~(\ref{eq:concept_loss}) is to trade off the contributions between likelihood and unlikelihood losses. We perform grid search using the TRECVid AVS datasets by varying $\alpha =\{0.0001,0.001,0.01,0.1,1\}$. The result is shown in Fig.~\ref{fig:alpha_parameter}. The best setting is $\alpha=0.01$ for all the search modes. We conduct a randomization test~\cite{randomization_test} to analyze the parameter sensitivity. The result shows no significant difference between the results for $\alpha$ being set as 0.1 and 0.01, respectively, at p-value $\leq 0.05$. We notice that when $\alpha=1$, the unlikelihood training tends to play a trick by assigning low probabilities to most exclusive concepts. Consequently, this results in locally exclusive concepts more likely being suppressed from decoding even if these concepts are mentioned in the captions.

\subsubsection{Sensitivity of $\theta$.} 
\label{Sec:sensitity_of_theta}

\begin{figure}[t]  
\centering  
\includegraphics[width=0.8\linewidth]{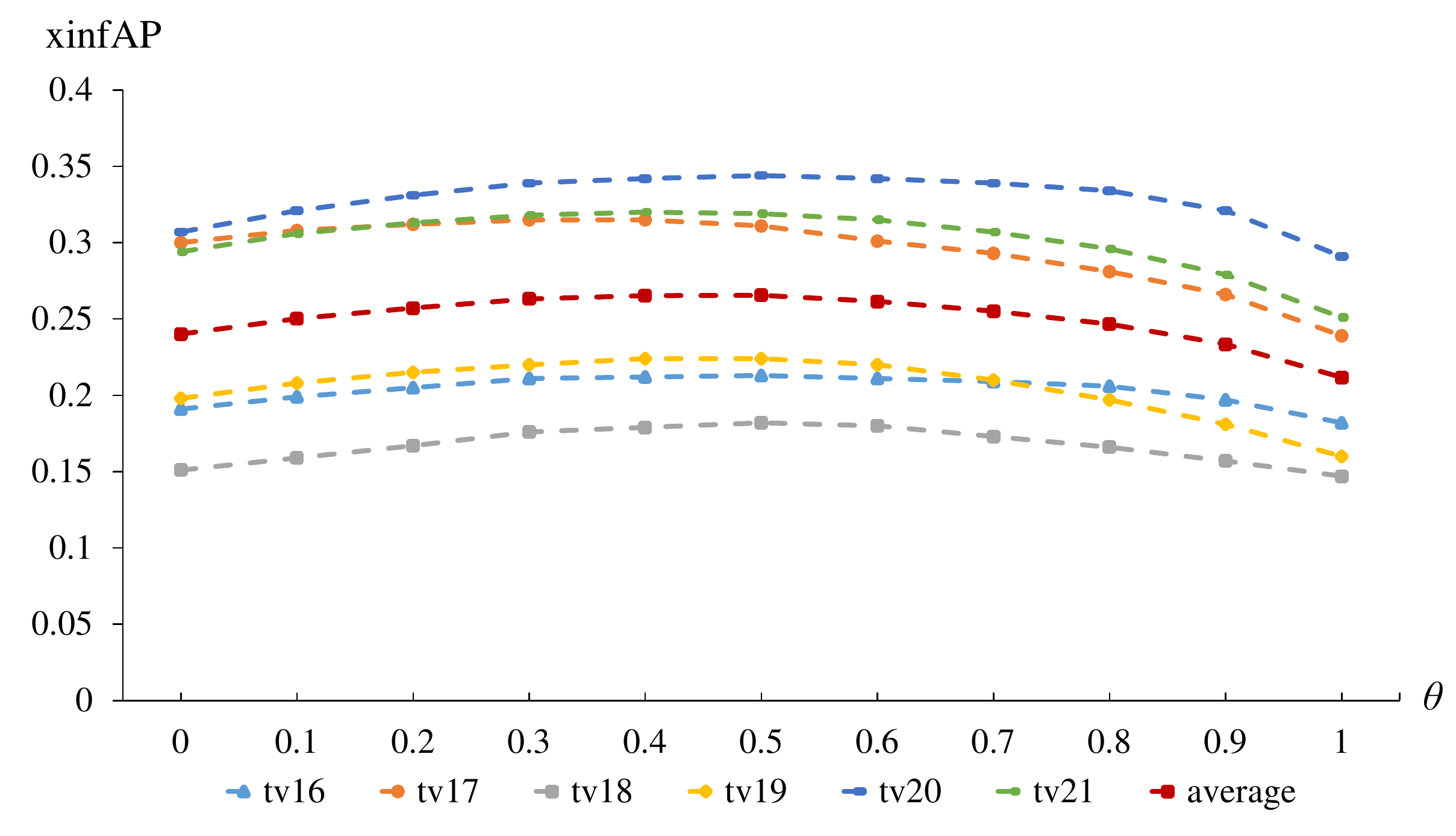}
\caption{Sensitivity of the hyper-parameter $\theta$ in fusion search on six query sets (tv16-tv21)} 
\label{fig:theta_parameter}
\end{figure}

The hyper-parameter $\theta$ in Eq. (\ref{eq:score_fusion}) controls the weights of concept-based search and concept-free search in fusion. Specifically, the extreme value $\theta=0$ represents concept-free only search (aka embedding search) while $\theta=1$ denotes concept-based only search. Fig.~\ref{fig:theta_parameter} shows the AVS performances with $\theta=\{$0, 0.1, 0.2, 0.3, 0.4, 0.5, 0.6, 0.7, 0.8, 0.9, 1.0$\}$ on the six query sets. As observed, the fused performances vary slightly in $[0.4,0.6]$. In the paper, we report $\theta=0.5$ as the fusion search.

\subsubsection{Video Representation}
\label{Sec:video_representation}

\begin{table*}[]
\caption{AVS performance comparisons (mean xinfAP) using different visual features across six TRECVid query sets (tv16-tv21).} 
\label{exp:different_feat}
\centering
\resizebox{\linewidth}{!}{
\begin{tabular}{l|cccccc}
\toprule
              & ResNet152+ResNext101
   & Swin Transformer    & SlowFast
 &
Concept-based Search   & Embedding Search   & Fusion Search    \\
              \hline
L1        & \checkmark    &     &     & 0.164      & 0.197 & 0.212  \\
L2        & \checkmark    & \checkmark &   & 0.180    &   0.220    & 0.235    \\
L3        & \checkmark    &     & \checkmark    &  0.199     & 0.222 & 0.244  \\
L4        & \checkmark    &  \checkmark   &  \checkmark   &  \textbf{0.209}     & \textbf{0.235} &  \textbf{0.261} \\

\bottomrule
\end{tabular}
}
\end{table*}

For the video representation, we extract three kinds of features in the paper. They are CNN features (resnet152 + resnext101 \cite{resnet,resnext}), vision transformer feature (Swin Transformer \cite{SwinTrans}) pre-trained on an image classification dataset (i.e., ImageNet \cite{imagenet}), and motion feature (SlowFast \cite{slowfast}) pre-trained on a video action dataset (i.e., Kinetic \cite{kinetics600}). Table \ref{exp:different_feat} contrasts the average AVS performances on six TRECVid query sets (tv16-tv21) using different combinations of these video features. The CNN features (i.e., resnet152+resNext101) serve as a baseline. As observed, the combination of all three kinds of features obtains the best result (i.e., L4). The model with diverse features (i.e., L3) beats the model with only visual features (i.e., L2). Specifically, the enhanced motion feature (i.e., SlowFast) improves the motion queries by detecting the target action more precisely. For example, for the query-653 \textit{Find shots of a group of people clapping}, the motion model obtains xinfAP~= 0.561 versus xinfAP = 0.111 of the visual-only models by having higher accuracy on the action clapping. Similarly, the enhanced visual feature (i.e., Swin Transformer) improves the baseline by having higher accuracy on visual items, such as the Christmas decoration in query-571 and the wheelchair in query-581. 

\subsection{Effects of CLIP and BLIP Features}
\label{Sec:clip_and_blip}

\begin{table}[t]
\caption{Performance comparison across different AVS models with and without CLIP or BLIP features.} 
\label{exp:comparison_clip}
\centering
\begin{tabular}{c ccc| ccc| c |c}

    \toprule
 & \multicolumn{3}{c|}{IACC.3} & \multicolumn{3}{c|}{V3C1} &V3C2 \\
                                                 \hline
  & tv16       & tv17      & tv18      & tv19            & tv20   & tv21  &tv22  &average    \\\hline
CLIP  \cite{CLIP}       & 0.182   &  0.205    & 0.082  &0.099   & 0.136  &  0.179  &0.123 &0.144 \\
BLIP-2 \cite{BLIP2}     &0.213 &0.226 & 0.168&0.199 &0.222 &0.273 &0.164 & 0.209 \\
\hline
w/o CLIP or BLIP-2 visual feature: & & & & & & & & \\
SEA* \cite{tmm2021-sea}             & 0.179         & 0.289           & 0.136       & 0.191     & 0.327  & 0.292 & 0.120 &0.219 \\ 
RIVRL* \cite{RIVRL}  &   0.137         &      0.212          &  0.108             &  0.145     &  0.248  &  0.195  &0.120 &0.166    \\
LAFF* \cite{LAFF}             &     0.173   &   0.263         &  0.138      &  0.175   &  0.270  & 0.255 &0.102  &0.197\\ 
ITV            &  0.211     &  0.292    &   \textbf{0.170}  &   \textbf{0.277}          &0.345  & 0.318  & 0.150 &0.252 \\
\hline

w/ CLIP visual feature: & & & & & & & & \\

SEA* \cite{tmm2021-sea}   & 0.200   &  \textbf{0.309}    & 0.140  &0.217  &0.344   &  0.330  &0.171 &0.244 \\
RIVRL* \cite{RIVRL}       & 0.152   &  0.235    & 0.116  &0.187  &0.278   &  0.234  &0.161 &0.195 \\
LAFF* \cite{LAFF}         & 0.170   &  0.279    & 0.137  &0.207  &0.295   &  0.280  &0.158 &0.218 \\

ITV           & 0.231   &   0.306  &  0.165   &  0.252         & 0.354    &   \textbf{0.380}  &0.184 &0.267 \\

\hline
w/ BLIP-2 visual feature: & & & & & & & & \\

SEA* \cite{tmm2021-sea}    &0.207 &0.292 &0.161 &0.219 &0.346 &0.338 &0.164 & 0.247 \\
RIVRL* \cite{RIVRL}        &0.163 &0.241 &0.130 &0.198 &0.304 & 0.262& 0.166& 0.209 \\
LAFF* \cite{LAFF}          &0.187 &0.256 &0.152 &0.206 &0.261 & 0.271& 0.146& 0.211\\

ITV            &\textbf{0.240} & 0.299 &0.168 & 0.256&\textbf{0.369} & 0.369& \textbf{0.187}& \textbf{0.270} \\

  \bottomrule  
\end{tabular}
\end{table}

Recently, the transformer pre-trained on large vision-language datasets, such as CLIP \cite{CLIP}, has been verified as a strong backbone to text-to-video retrieval \cite{HTW:vibro_vbs2023,visione:vbs2023,Luo2021CLIP4Clip}. For example, CLIP features have been shown to improve the precision of search in Video Browser Showdown 2023 \cite{HTW:vibro_vbs2023}. Variants of the CLIP model (e.g., CLIP4CLIP \cite{Luo2021CLIP4Clip}) have also demonstrated new state-of-the-art results, such as on MSR-VTT \cite{msr-vtt} and VATEX \cite{VATEX} datasets. In addition to CLIP, BLIP-2 \cite{BLIP2} is also popularly used.

In this section, we investigate the integration of CLIP (as well as BLIP-2) features along with other existing features in ITV for learning interpretable embeddings. Specifically, the visual encoders of these pre-trained models are employed to extract video features. Together with other features stated in Section \ref{method_search_model}, the encoders and the decoder of ITV are trained.

The result is shown in Table \ref{exp:comparison_clip}, which contrasts the search performances after employing these features for model training. For a fair comparison, all the compared methods are trained in a similar fashion with the visual features of CLIP or BLIP-2. In addition, we include the baselines, i.e., CLIP and BLIP-2, which directly rank video shots by comparing their textual and visual features using cosine similarity. Employing these features generally boosts the search performances for all the compared methods. Specifically, for CLIP, ITV shows higher xinfAP scores on 113 out of 190 queries, with corresponding numbers of 118, 107, and 116 for SEA, RIVRL, and LAFF, respectively. Although the baseline CLIP does not perform well, there are specific queries (e.g., query-589 \textit{Find shots of car driving scenes in a rainy day}) with terms (e.g., rainy day) that CLIP either performs better or boosts the performances of other approaches. BLIP-2 exhibits higher performance than CLIP. ITV, SEA, and RIVRL show slightly better average performance with BLIP-2 than CLIP. Such improvement, nevertheless, is not noticed for LAFF. In general, for vocabularies that are absent or rare in ITV (e.g., destroyed buildings), employing the CLIP or BLIP-2 feature in ITV is helpful in search performance. Similar observations are also noted for other approaches. While all the approaches show noticeable improvements, ITV still demonstrates the best performance on average. This is mainly due to the ability of ITV to decode coherent concepts for the interpretation of embedding. Other approaches remain suffering from not being able to differentiate exclusive concept pairs (e.g., \textit{indoor} versus \textit{outdoor}, \textit{black} versus \textit{white}) despite with CLIP or BLIP-2 features. Their performances (with CLIP or BLIP-2) are indeed not better than ITV (without CLIP or BLIP-2).

\begin{figure*}[t]  
\centering  
\includegraphics[width=1\linewidth]{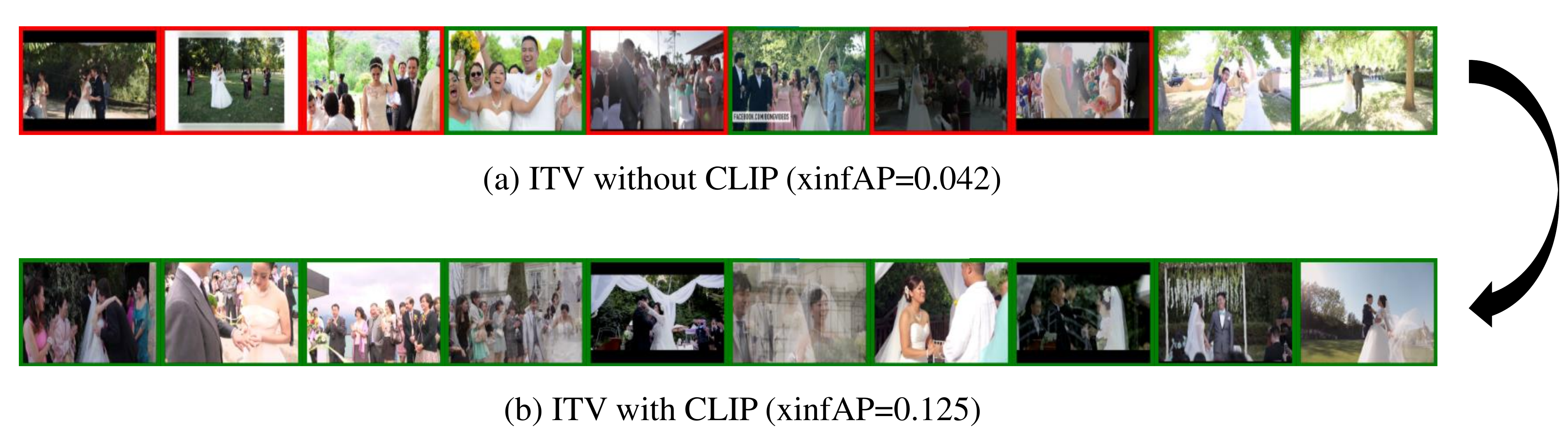}
\caption{Performances before and after integrating with the CLIP feature on the query-715 \textit{Find shots of Asian bride and groom celebrating outdoors}.} 
\label{fig:asian_bride_and_groom}
\end{figure*} 

\begin{figure*}[t]  
\centering  
\includegraphics[width=1\linewidth]{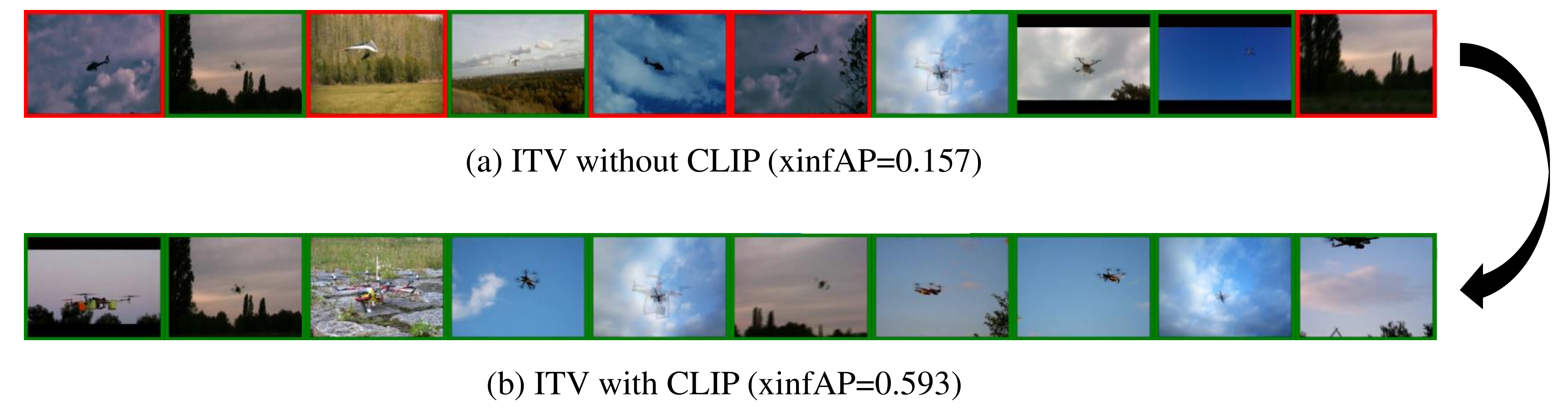}
\caption{Performances before and after integrating with the CLIP feature on the query-611 \textit{Find shots of a drone flying}.} 
\label{fig:drone_flying}
\end{figure*} 

Figures \ref{fig:asian_bride_and_groom}-\ref{fig:drone_flying} show examples demonstrating how the CLIP feature boosts the ITV performance. In Fig. \ref{fig:asian_bride_and_groom}, the term \textit{Asian bride and groom} is absent in the vocabulary, and therefore, ITV finds a mixture of non-Asian and Asian couples. With the CLIP feature, ITV is able to rank the top 10 with correct video shots. In Fig. \ref{fig:drone_flying}, ITV cannot distinguish helicopter and plane from drone as the \textit{drone} is a rare concept seldom annotated in the training examples. With CLIP features as input to ITV, the issue due to vocabulary imbalance is alleviated.

\section{Conclusion}
\label{Sec:conclusion}
We have presented (un)likelihood training for cross-modal representation learning. The experiments on TRECVid datasets, particularly, verify the merit of having semantically consistent interpretation for both video and query embeddings to boost text-to-video search performance. The proposed training objectives effectively address the problem of sparse training labels in learning cross-modal embedding. While likelihood training tends to densely predict more words beyond those labeled in the training examples for interpretation, the unlikelihood objective prunes inconsistent concepts to ensure semantic coherence. From the experiment analysis, the prediction bias towards frequent, co-occurring, and contradicting concepts can be alleviated by the unlikelihood objective, resulting in higher interpretation quality for a better chance of having rare and interesting concepts. This is especially evidential for query embedding, where novel terms are expanded for query as interpretation, leading to higher and robust search performance. The encouraging result of unlikelihood training, however, is at the expense of occasionally removing relevant concepts. While this side effect impacts concept-based search, the performance of embedding search remains noticeably stable. Overall, the best performances are consistently observed in most queries for training with both objective functions and for inference with fusion search that leverages embeddings as well as their interpreted concepts. Furthermore, the (un)likelihood training also benefits from more powerful features (e.g., CLIP, BLIP-2) being included for learning interpretable embeddings.

\begin{acks}
This research was supported by the Singapore Ministry of Education (MOE) Academic Research Fund (AcRF) Tier 1 grant and the CityU MF\_EXT (project no. 9678180).
\end{acks}



\bibliographystyle{ACM-Reference-Format}
\bibliography{UL}
\end{document}